\documentclass[runningheads]{llncs}
\usepackage{graphicx}
\usepackage{amsmath,amssymb} % define this before the line numbering.
\usepackage{mathtools}

\usepackage{color}
\usepackage[width=122mm,left=12mm,paperwidth=146mm,height=193mm,top=12mm,paperheight=217mm]{geometry}
\usepackage{multirow}
\usepackage{float}
\usepackage{adjustbox}
\usepackage{subcaption}
\begin{document}

\title{Deep Continuous Fusion for Multi-Sensor 3D Object Detection}
\titlerunning{Deep Continuous Fusion}

\author{Ming Liang\inst{1} \and
Bin Yang\inst{1,2} \and
Shenlong Wang\inst{1,2} \and
Raquel Urtasun\inst{1,2}}
\authorrunning{M. Liang, B. Yang, S. Wang and R. Urtasun}

\institute{Uber Advanced Technologies Group \and 
University of Toronto \\
\email{\{ming.liang, byang10, slwang, urtasun\}@uber.com}}

\maketitle

%!TEX root = 2683.tex
\begin{abstract}

In this paper, we propose a novel 3D object detector  that can exploit both LIDAR as well as cameras to perform very accurate localization. Towards this goal, we design an end-to-end learnable architecture that exploits continuous convolutions to fuse image and LIDAR feature maps at different levels of resolution. Our proposed continuous fusion layer encode both discrete-state image features as well as continuous geometric information. This enables us to design a novel, reliable and efficient end-to-end learnable 3D object detector based on multiple sensors. Our experimental evaluation on both KITTI as well as a large scale 3D object detection benchmark shows significant improvements over the state of the art.

\end{abstract}

%!TEX root = top.tex
\section{Introduction}

One of the fundamental problems when building perception systems for autonomous driving is to be able to detect objects in 3D space. An autonomous vehicle needs to perceive the  objects present in the  3D scene from its sensors in order to plan its motion  safely. Most self-driving vehicles are equipped with both cameras and 3D sensors. This brings  potential to exploit the advantage of different sensor modalities to conduct accurate and reliable 3D object detection. 
During the past years, 2D object detection from camera images has seen significant progress \cite{rcnn,frcn,faster,rfcn,fpn,yolo,ssd,focal}. However, there is still large space for improvement when it comes to object localization in 3D space. 

LIDAR based 3D object detection has drawn much attention recently when combined with the power of deep learning. Representative works either project the 3D LIDAR points onto camera perspective \cite{velofcn}, overhead view \cite{dpt,pixor,3dfcn,DoBEM}, 3D volumes \cite{voxelnet,vote3deep} or directly conduct 3D bounding box estimation over unordered 3D points \cite{fpointnet}. However,  these approaches suffer at long range and when dealing with occluded objects due to the sparsity of the LIDAR returns over these regions.

Images, on the other hand, provide dense measurements, but precise 3D localization is hard  due to the loss of depth information caused by perspective projection, particularly when using monocular cameras \cite{mono3d,3doppami}. Recently, several approaches have tried to exploit both cameras and LIDAR jointly. In \cite{fpointnet,3dop} camera view is used to generate proposals while LIDAR is used to conduct the final 3D localization. However, these cascading approaches do not exploit the capability to perform joint reasoning over multi-sensor's inputs. As a consequence, the 3D detection performance is bounded by the  2D image-only detection step. Other approaches \cite{mv3d,avod,fpccnn} apply 2D convolutional networks on both camera image and LIDAR bird's eye view (BEV) representations, and fuse them at the intermediate region-wise convolutional feature map via feature concatenation. This fusion usually happens at a  coarse level, with significant resolution loss. Thus, it remains an open problem to design 3D detectors that can  better exploit multiple modalities.  The challenge lies in the fact that the LIDAR points are sparse and continuous, while cameras capture dense features at discrete state; thus, fusing them is non-trivial. 

In this paper, we propose a 3D object detector that reasons in bird's eye view (BEV) and fuses image features by learning to project them into BEV space. Towards this goal, we design an end-to-end learnable architecture that exploits continuous convolutions to fuse image and LIDAR feature maps at different levels of resolution. The proposed continuous fusion layer is capable of encoding dense accurate geometric relationships between positions under the two modalities. This enables us to design a novel, reliable and efficient 3D object detector based on multiple sensors. Our experimental evaluation on both KITTI \cite{kitti} and a large scale 3D object detection benchmark \cite{pixor} shows significant improvements over the state of the art. 

%!TEX root = top.tex
\section{Related Work}

\paragraph{LIDAR-Based Detection:} Several detectors have been proposed recently to produce accurate localization from 3D sensors. VeloFCN \cite{velofcn} projects the LIDAR points  to front view and applies a 2D fully convolutional network on the front-view representation to generate 3D detections. 3DFCN \cite{3dfcn} exploits a bird's eye view  representation of the LIDAR and applies a 3D fully convolutional network. PIXOR \cite{pixor} conducts a single-stage, proposal-free detection over a height-encoded bird's eye view representation. FAF \cite{dpt} conducts detection, tracking and short-term future prediction jointly within a single network.

\paragraph{Joint Camera-3D Sensor Detection:} Over the past few years, many techniques have explored both cameras and 3D sensors jointly to perform 3D reasoning. One common practice is to perform depth image based processing, which encodes the 3D geometry as an additional image channel \cite{silberman2012indoor,rcnn,gupta2014learning}. For instance,  \cite{gupta2014learning} proposes a novel geocentric embedding for the depth image and, through concatenating with the RGB image features, significant improvement can be achieved. However, the output space of these approaches is on the camera image plane. In the context of autonomous driving, this is not desirable as we wish to localize objects in 3D space. 
Additional efforts have to be made in order to generate amodal object bounding boxes in 3D. An alternative idea is to use voxelization, and consider the color image in the voxels as additional channels \cite{ss,dss}. However, this is not efficient in memory and computation, and color information is lost over many voxels due to the perspective projection. Other methods exploit one sensor modality to generate bounding box proposals and another to conduct final classification and regression. For instance, \cite{3dop} exploits a depth map to generate 3D object proposals and uses images to perform box classification. On the other hand, F-PointNet \cite{fpointnet} uses the camera to generate 2D proposals, and PointNet \cite{pointnet} to directly predict the 3D shape and location within the visual frustum produced by 2D bounding box. \cite{mv3d} proposes to fuse features from multiple sensors in multiple views through ROI-pooling. However, accurate geometric information is lost in this coarse-level region-based pooling scheme.

\paragraph{Convolution on 3D Point Clouds:} Our approach is also related to the line of work that conducts learnable convolution-like operators over point clouds. Graph (convolutional) neural networks \cite{gnn,gcn,geometric-deep-learning} consider each point as a node in the graph while the edges are built through spatial proximity. Messages are sent between nodes to propagate information. Another family of methods designs the convolution  \cite{pccn,ecc,monet,anisotropic-cnn} or pooling operators \cite{pointnet,pointnet2} directly over points or 3D meshes. These approaches are more powerful and able to encode geometric relationship without losing accuracy. Our proposed continuous fusion layer can be considered as a special case that connects points between different modalities. 

%!TEX root = top.tex
\section{Multi-sensor 3D Object Detection}

Recently, several works \cite{3dfcn,DoBEM,voxelnet,avod,mv3d,3doppami} have shown very promising results  by performing  3D object detection in BEV. These detectors are effective as BEV maintains  the  structure native to 3D sensors such as LIDAR. As a consequence, convolutional networks can be easily trained and  strong priors like object size can be exploited. 
Since most self-driving cars are equipped with both LIDAR and cameras, sensor fusion between these modalities is desirable in order to further boost performance. 

\begin{figure*}[t]
	\centering
	\includegraphics[width=0.998\textwidth]{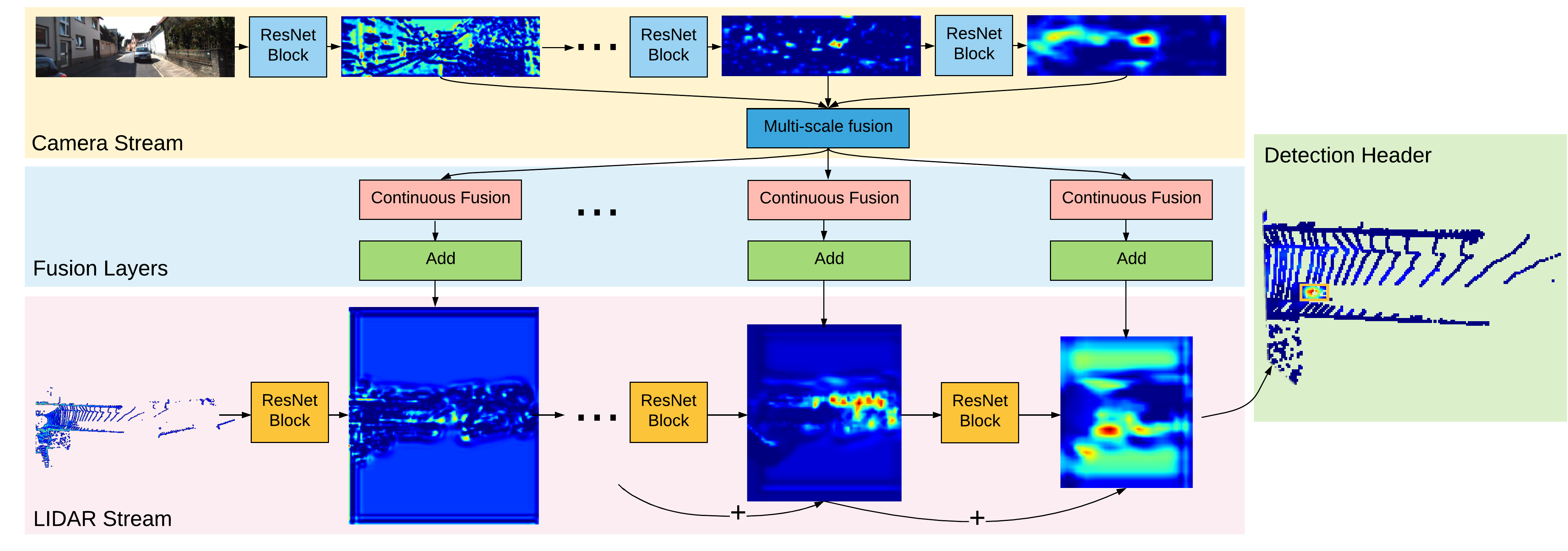}
	\caption{Architecture of our model. There are two streams, namely the camera image stream and the BEV LIDAR stream. Continuous fusion layers are used to fuse the image features onto the BEV feature maps.}
	\label{fig:architecture}
\end{figure*}

Fusing information between LIDAR and images is non-trivial as images represent a projection of the world onto the camera plane, while LIDAR captures the world's native 3D structure. 
One possibility is to project the LIDAR points onto the image, append an extra channel with depth information and exploit traditional 2D detection architectures. This has been shown to be very effective when reasoning in image space (e.g., \cite{silberman2012indoor,gupta2014learning,eitel2015multimodal}).
Unfortunately, a second step is necessary in order to obtain 3D detections from the 2D outputs. 

In contrast, in this paper we perform the opposite operation. We exploit image features  extracted by a convolutional network, and then project the image features into BEV and fuse them with the convolution layers of a LIDAR based detector. 
This fusing operation is non-trivial, as image features happen at discrete locations; thus, one needs to ``interpolate" to create a dense BEV feature map. To perform this operation, we take advantage of continuous convolutions \cite{pccn} to extract information from the nearest corresponding image features for each point in BEV space. Our overall architecture includes two streams, with one stream extracting image features and another one extracting features from LIDAR BEV. 
We design the continuous fusion layer to bridge multiple intermediate layers on both sides in order to perform multi-sensor fusion at multiple scales. This architecture allows us to  generate the final detection results in BEV space, as desired by our autonomous driving application. We refer the reader to Fig. \ref{fig:architecture} for an illustration of our architecture. 

In the remainder of the section, we first review continuous convolutions, and then show how they can be exploited to fuse information from LIDAR and images. After that we propose a deep multi-sensor detection architecture using this new continuous fusion layer.

\subsection{Continuous Fusion Layer}
\paragraph{Deep Parametric Continuous Convolution:} Deep parametric continuous convolution \cite{pccn} is a learnable operator that operates over non-grid-structured data. The motivation behind this operator is to extend the standard grid-structured convolution to non-grid-structured data, while retaining high capacity and low complexity. The key idea is to exploit multi-layer perceptron (MLP) as parameterized kernel functions for continuous convolution. This parametric kernel function spans the full continuous domain. Furthermore, the weighted summation over finite number of neighboring points is used to approximate the otherwise computationally prohibitive continuous convolution.  Each neighbor is weighted differently according to its relative geometric offset with regard to the target point. More specifically, parametric continuous convolution conducts the following operation: \[\mathbf{h}_i = \sum_j  \mathrm{MLP} (\mathbf{x}_i - \mathbf{x}_j) \cdot \mathbf{f}_j\] where $j$ indexes over the neighbors of point $i$, $\mathbf{f}_j$ is the input feature and $\mathbf{x}_j$ is the continuous coordinate associated with a point. The MLP computes the convolutional weight at each neighbor point. The advantage of parametric continuous convolution is that it utilizes the concept of standard convolution to capture local information from neighboring observations, without a rasterization stage that could lead to geometric information loss. In this paper we argue that continuous convolution is a good fit for our task, due to the fact that both camera view and BEV are connected through a 3D point set, and modeling such  geometric relationships between them in a lossless manner is key to fusing information. 

\begin{figure*}[t]
	\centering
	\includegraphics[width=0.85\textwidth]{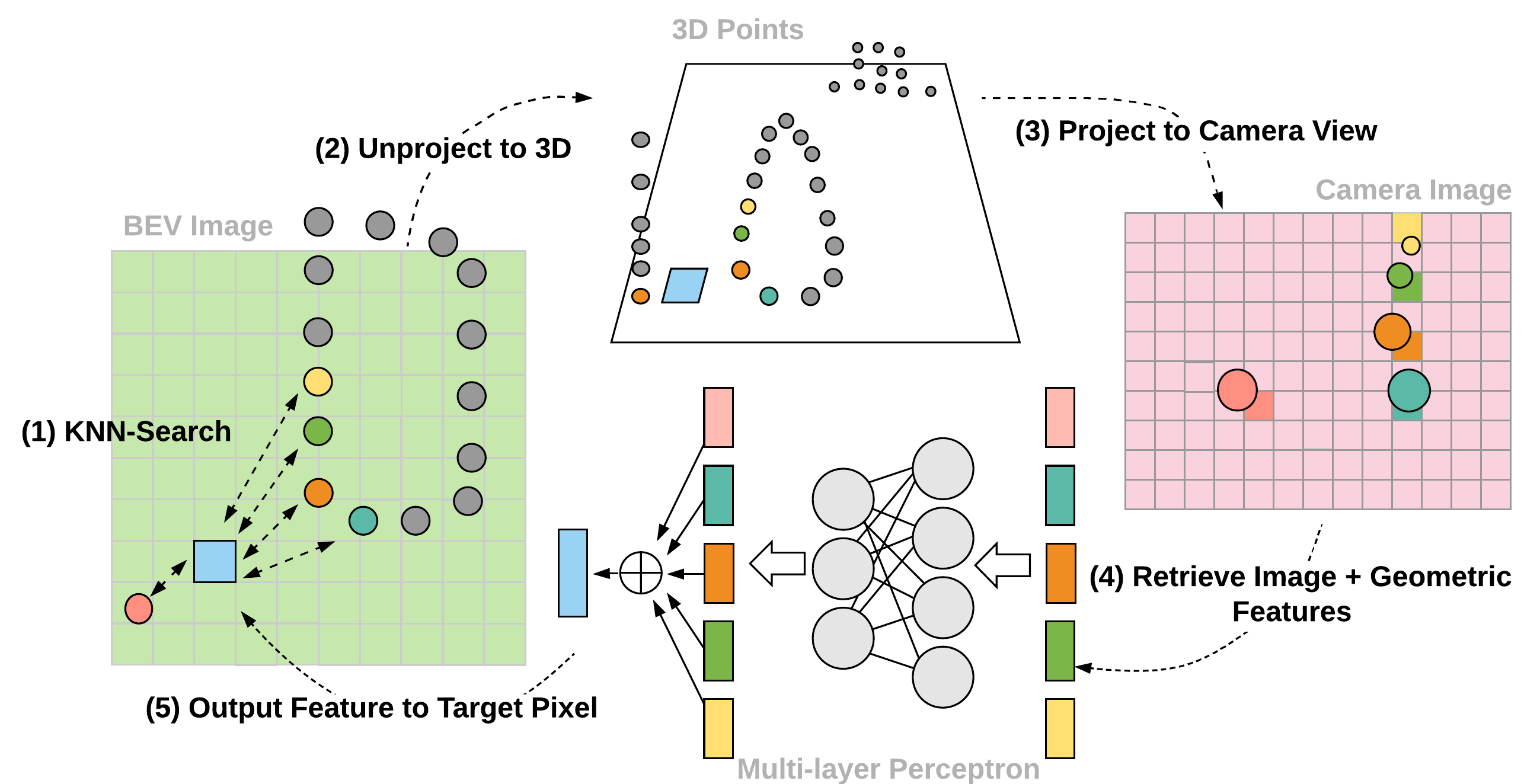}
	\caption{Continuous fusion layer: given a target pixel on BEV image, we first extract K nearest LIDAR points (Step 1); we then project the 3D points onto the camera image plane (Step 2-Step 3); this helps retrieve corresponding image features (Step 4); finally we feed the image feature + continuous geometry offset into a MLP to generate feature for the target pixel (Step 5). }
	\label{continuous-fusion}
\end{figure*}

\paragraph{Continuous Fusion Layer:} Our proposed continuous fusion layer exploits continuous convolutions to overcome the two aforementioned problems, namely the sparsity in the observations and the handling of the spatially-discrete features in camera view image. Given the input camera image feature map and a set of LIDAR points, the target of the continuous fusion layer is to create a dense BEV feature map where each discrete pixel contains features generated from the camera image. This dense feature map can then be readily fused with BEV feature maps extracted from LIDAR. One difficulty of image-BEV fusion is that not all the discrete pixels on BEV space are observable in the camera. To overcome this, for each target pixel in the dense map, we find its nearest K LIDAR points over the 2D BEV plane using Euclidean distance.  We then exploit MLP to fuse information from these K nearest points to ``interpolate'' the unobserved feature at the target pixel. 
For each source LIDAR point, the input of our MLP contains two parts:  First, we extract the corresponding image features by projecting the source LIDAR point onto the image plane. Bilinear interpolation is used to get the image feature at the continuous coordinates.
Second, we encode the 3D neighboring offset between the source LIDAR point and the target pixel on the dense BEV feature map, in order to model the dependence of each LIDAR point's contribution on its relative position to the target. Overall, this gives us a $K\times(D_i+3)$-d input to the MLP for each target pixel, where $D_i$ is the input feature dimension.  For each target pixel, the MLP outputs a $D_o$-dimensional output feature by summing over the MLP output for all its neighbors.

That is to say:
\[\mathbf{h}_i = \sum_j \mathrm{MLP}(\mathrm{concat}\left[\mathbf{f}_j, \mathbf{x}_j-\mathbf{x}_i \right])\]
where $\mathbf{f}_j$ is the input image feature of point $j$, $\mathbf{x}_j - \mathbf{x}_i$ is the 3D offset from neighbor point $j$ to the target $i$ and $\mathrm{concat}(\cdot)$ is the concatenation of multiple vectors.
In practice, we utilize a 3-layer perceptron where each layer has $D_i$ hidden features.  The MLP's output features are then combined by element-wise summation with the BEV features from the previous layer to fuse multi-sensor information. The overall computation graph is shown in 
Fig.~\ref{continuous-fusion}. 
\paragraph{Comparison against Standard Continuous Convolution:} Compared against standard parametric continuous convolution \cite{pccn}, the proposed continuous fusion layer utilizes MLP to directly output the target feature, instead of outputting weights to sum over features. This gives us stronger capability and more flexibility to aggregate information from multiple neighbors. Another advantage is memory-efficiency. Since the MLP directly outputs features rather than weights, our approach does not need to explicitly store an additional weighting matrix in GPU memory. 

\subsection{Multi-Sensor Object Detection Network} 
Our multi-sensor detection network has two streams: the image feature network and the  BEV network. We use four continuous fusion layers to fuse multiple scales of image features into BEV network from lower level to higher level. The overall architecture is depicted in Fig.~\ref{fig:architecture}. In this section we will discuss each individual component in more details. 
\paragraph{Backbone Networks:}
We choose the lightweight ResNet18 as the backbone of the image network because of its efficiency. In our application domain, real time estimates are crucial for safety. 
The BEV network is customized to have five groups of residual blocks. 
The number of convolution layers per each group is 2, 4, 8, 12 and 12 respectively. All groups start with a stride 2 convolution except for the first group, and all other convolutions have stride 1.
The feature dimension of each group is 32, 64, 128, 192 and 256 respectively. 

\paragraph{Fusion Layers:}
Four continuous fusion layers are used to fuse multi-scale image features into the four residual groups of the BEV network. The input of each continuous fuse layer is an image feature map combined from the outputs of all four image residual groups. We use the same combination approach as the feature pyramid network \cite{fpn}. The output feature in BEV space has the same shape as the corresponding BEV layer and is combined into BEV through element-wise addition. Our final BEV feature output also combines the last three residual groups' output in a similar manner as FPN \cite{fpn}, in order to exploit multi-scale information.

\paragraph{Detection Header:} We use a simple detection header for real-time efficiency. A $1\times 1$ convolutional layer is computed over the final BEV layer to generate the detection output. At each output location we use two anchors which have fixed size and two orientations, $0$ and $\pi / 2$ radians respectively. Each anchor's output includes the per-pixel class confidence and its associated box's center location, size and orientation. A Non-Maximum Suppression (NMS) layer follows to generate the final object boxes based on the output map.

\paragraph{Training:} 
We use a multi-task loss to train our network.
Following common practice in object detection  \cite{rcnn,frcn,faster}, we define the loss function as the summation of classification and regression losses.
	\begin{equation}
	L = L_{cls} + \alpha L_{reg}
	\end{equation}
	where $L_{cls}$ and $L_{reg}$ are the classification loss and regression loss, respectively. $L_{cls}$ is defined as the binary cross entropy between class confidence and  the label
	\begin{equation}
	L_{cls} = \frac{1}{N}\left(l_c\log(p_c) + (1-l_c)\log{(1-p_c)}\right)
	\end{equation}
	where $p_c$ is the predicted classification score, $l_c$ is the binary label, and $N$ is the number of samples. For 3D detection, $L_{reg}$ is the sum of seven terms
	\begin{equation}\label{equ:reg}
	L_{reg} = \frac{1}{N_{pos}}\sum_{k\in(x, y, z, w, h, d, t)}{D(p_k, l_k)}
	\end{equation}
	where $(x, y, z)$ denotes the 3D box center, $(w, h, d)$ denotes the box size, $t$ denotes the orientation, and $N_{pos}$ is the number of positive samples. $D$ is a smoothed L1-norm defined as:
	\begin{equation}
	D(p_k, l_k) =
	\begin{cases}
	0.5(p_k-l_k)^2& \text{if } |p_k-l_k| < 1\\
	|p_k-l_k| - 0.5& \text{otherwise},
	\end{cases}
	\end{equation}
with	$p_k$ and $l_k$ the predicted and ground truth offsets respectively. For $k\in{(x, y, z)}$, $p_k$ is encoded as:
	\begin{equation}
	p_k = (k - a_k)/a_k
	\end{equation}
	where $a_k$ is the coordinate of the anchor. For $k\in{(w, h, d)}$, $p_k$ is encoded as:
	\begin{equation}
	p_k=\log(k/a_k)
	\end{equation}
	where $a_k$ is the size of anchor. The orientation offset is simply defined as the difference between predicted and labeled orientations:
	\begin{equation}
	p_t = k-a_k
	\end{equation}

When only BEV detections are required, the $z$ and $d$ terms are removed from the regression loss.
Positive and negative samples are determined based on distance to the ground-truth object center. Hard negative mining is used to sample the negatives. In particular, we first randomly select $5\%$ negative anchors and then only use top-$k$ among them for training, based on the classification score.
We initialize the image network with ImageNet pre-trained weights and initialize the BEV network and continuous fusion layers using Xavier initialization \cite{xavier}. 
The whole network is trained end-to-end through back-propagation. Note that there is no direct supervision on the image stream; instead, error is propagated along the bridge of continuous fusion layer from the BEV feature space.

%!TEX root = top.tex

%=================== kitti-evaluation-results ================
\begin{table*}[t]
\begin{center}
%\begin{small}
\addtolength{\tabcolsep}{-0pt}
\begin{tabular}{|c|c|c|ccc|ccc|}
\hline
\multirow{2}{*}{Method}& \multirow{2}{*}{Input}& \multirow{2}{*}{Time (s)}& \multicolumn{3}{|c|}{3D AP (\%)} & \multicolumn{3}{|c|}{BEV AP (\%)}\\
\cline{4-9}
 & & & easy & moderate & hard & easy & moderate & hard \\
\hline
MV3D~\cite{mv3d} & LIDAR & 0.24 & 66.77 & 52.73 & 51.31 & 85.82 & 77.00 & 68.94 \\
VxNet~\cite{voxelnet} & LIDAR & 0.22 & 77.49 & 65.11 & 57.73 & \bf 89.35 & 79.26 & 77.39 \\
NVLidarNet & LIDAR & 0.1 & n/a & n/a & n/a & 84.44 & 80.04 & 74.31 \\
PIXOR~\cite{pixor} & LIDAR & 0.035 & n/a & n/a & n/a & 87.25 & 81.92 & 76.01 \\
\hline
F-PC\_CNN~\cite{fpccnn} & LIDAR+Img & 0.5 & 60.06 & 48.07 & 45.22 & 83.77 & 75.26 & 70.17 \\
MV3D~\cite{mv3d} & LIDAR+Img & 0.36 & 71.09 & 62.35 & 55.12 & 86.02 & 76.90 & 68.49 \\
AVOD-FPN~\cite{avod} & LIDAR+Img & 0.1 & 81.94 & \bf 71.88 & \bf 66.38 & 88.53 & 83.79 & 77.90\\
F-PointNet~\cite{fpointnet} & LIDAR+Img & 0.17 & 81.20 & 70.39 & 62.19 & 88.70 & 84.00 & 75.33 \\
AVOD~\cite{avod} & LIDAR+Img & 0.08 & 73.59 & 65.78 & 58.38 & 86.80 & 85.44 & \bf 77.73 \\
\hline
Our Cont Fuse & LIDAR+Img & {0.06} & {\bf 82.54} & {66.22} & {64.04} & {88.81} & {\bf 85.83} & {77.33} \\
\hline
\end{tabular}
%\end{small}
\caption{Evaluation on KITTI 3D and Bird's-Eye-View (BEV) Object Detection Benchmark (Car).}
\label{tab:kitti_number}
\end{center}
\end{table*}
%==============================================================

\section{Experiments}
We evaluate our multi-sensor 3D object detector on two datasets: the public KITTI benchmark \cite{kitti} and a large-scale 3D object detection dataset (TOR4D) \cite{pixor}. On the public KITTI dataset we compare with other state-of-the-art methods in both 3D object detection and BEV object detection tasks. An ablation study is also conducted that compares different model design choices. We also evaluate our model on TOR4D, a large-scale 3D object detection dataset collected in-house on roads of North-American cities. On this dataset we show that the proposed approach works particularly well in long-range ($>60$m) detection, which plays an important role in practical object detection systems for autonomous driving. Finally we show qualitative results and discuss future directions.

\subsection{KITTI 3D/BEV Object Detection}
\subsubsection{Dataset and metric}
KITTI \cite{kitti} object detection benchmark has 7,481 training frames and 7,518 testing frames. For each frame, an RGB camera image is shot by a front-facing camera mounted on top of the vehicle, and a 3D LIDAR point cloud is captured by a laser scanner (Velodyne HDL-64E) mounted on top of the vehicle. KITTI annotates objects that appear in camera view with ground-aligned 3D bounding boxes. For detection, we evaluate on ``Car'' class only because other classes do not have enough labels to train our neural network based method.

Detection results on the testing set are submitted to the KITTI evaluation server for evaluation. 
The 11-point AP is used as the official metric. For 3D object detection task, 3D Intersection-Over-Union (IoU) is used to distinguish between true positive and false positive with a threshold of 0.7. For BEV object detection, 2D IoU on BEV is used with the same threshold. ``DontCare'' and ``Van'' classes do not count as false positives. KITTI divides the labels into three subsets: easy, moderate and hard, according to the heights of their 2D bounding boxes, occlusion levels and truncation levels. The leaderboard ranks all entries by AP in the moderate subset.

\subsubsection{Implementation details}
All camera images are cropped to the size of $370\times 1224$. To generate the BEV input, the 3D space is voxelized into a $512 \times 448 \times 32$ volume, corresponding to 70 meters in front direction, $\pm$40 meters on left and right sides of the ego-car. 8-neighbor interpolation is used during voxelization. We train a 3D multi-sensor fusion detection model, where all seven regression terms in Equation \ref{equ:reg} are used. Because the height of the 2D box is needed by KITTI evaluation server, we add another regression term to predict the 2D height. As a result, we have a final output tensor with the size $118 \times 112 \times 2 \times 9$, where $118\times112$ is the number of spatial anchors and 2 is the number of orientation anchors.

Since the training data in KITTI is limited, we adopt several data augmentation techniques to alleviate over-fitting. For each frame during training, we apply random scaling ($0.9\sim1.1$ for all 3 axes), translation ($-5\sim5$ meters for $xy$ axes and $-1\sim1$ for $z$ axis) and rotation ($-5\sim5$ degrees along $z$ axis) on 3D LIDAR point clouds, and random scaling ($0.9\sim1.1$) and translation ($-50\sim50$ pixels) on camera images. We modify the transformation matrix from LIDAR to camera accordingly to ensure their correspondence. We do not apply data augmentation during testing.

We train the model with a batch size of 16 on 4 GPUs. Adam \cite{adam} is used for optimization with 0 weight decay. The learning rate is initialized as 0.001, and later decayed by 0.1 at 30 epochs and 45 epochs. The training ends after 50 epochs.

\subsubsection{Evaluation results}
We compare our 3D detector with other state-of-the-art methods in Table \ref{tab:kitti_number}. We divide all comparing methods into two categories depending on whether the image is used. For BEV detection, our model outperforms all other methods (measured by moderate AP). For 3D detection, our model ranks third among the models, but has the best AP on the easy subset. While keeping a high detection accuracy, our model is able to run at real-time efficiency. Our detector runs at $>15$ frames per second, much faster than all other LIDAR based and fusion based methods.

%=================== kitti-evaluation-results ================
\begin{table*}[t]
	\begin{center}
		%\begin{small}
		\addtolength{\tabcolsep}{-0pt}
		\begin{tabular}{|c|c|c|c|ccc|ccc|}
			\hline
			\multirow{2}{*}{Input} & KNN & Geometric & KNN & \multicolumn{3}{|c|}{3D AP (\%)} & \multicolumn{3}{|c|}{BEV AP (\%)}\\
			 & pooling & feature & parameter & easy & mod. & hard & easy & mod. & hard \\
			\hline
			LIDAR & n/a & n/a & n/a & 78.08 & 65.90 & 61.51 & 92.16 & 84.23 & 80.40\\
			\hline
			\multirow{6}{*}{\shortstack{LIDAR\\+IMG}} & no & no & n/a & 81.50 & 67.79 & 63.05 & 92.10 & 85.18 & 81.85\\
			\cline{2-10}
			 & \multirow{5}{*}{yes} & no & k=1, d=10 & 81.93 & 70.09 & 65.63 & 93.50 & 86.37 & 81.73\\ 
			\cline{3-10}
			 &  & \multirow{4}{*}{yes} & k=1, d=3 & 84.58 & 72.33 & 67.50 & 93.84 & 86.10 & 82.00\\
			 &  &  & k=5, d=3 & 82.58 & 71.20 & 66.52 & 93.53 & 86.79 & 81.97\\
			 &  &  & k=1, d=10 & \bf 86.32 & \bf 73.25 & \bf 67.81 &  95.44 & 87.34 & \bf 82.43\\
			 &  &  & k=1, d=+inf & 86.29 & 72.39 & 67.45 & \bf 95.90 & \bf 87.39 & 82.41\\
			\hline
		\end{tabular}
		%\end{small}
		\caption{Ablation Study on KITTI 3D and Bird's-Eye-View (BEV) Object Detection Benchmark (Car). We compare our continuous fusion model with a LIDAR only model (LIDAR input), a sparse fusion model (no KNN pooling) and a discrete fusion model (no geometric feature).}
		\label{tab:ablation}
	\end{center}
\end{table*}
%==============================================================

\subsection{Ablation Study on KITTI}
Continuous fusion has two components which enables the dense accurate fusion between image and LIDAR. The first is KNN pooling, which gathers image feature input for dense BEV pixels through sparse neighboring points. The second is the geometric feature input to MLP, which compensates for the continuous offsets between the matched position pairs of the two modalities. We investigate these components by comparing the continuous fusion model with a set of derived models. We also investigate the model with different KNN hyper-parameters. The experiments are conducted on the same training/validation split provided by MV3D \cite{mv3d}. We modify KITTI's AP metric from 11-point area-under-curve to 100-point for smaller variance. In practice, these two versions have $<1\%$ discrepancy.

The first derived model is a LIDAR BEV only model, which uses the BEV stream of the continuous fusion model as its backbone net and the same detection header. All continuous fusion models significantly outperform the BEV model in all six metrics, which demonstrates the great advantage of our model. This advantage is even larger for 3D detection, suggesting that the fused image features provide complementary $z$ axis information to BEV features.

The second derived model is a discrete fusion model, which has neither KNN pooling nor geometric feature. This model projects the LIDAR points onto image and BEV to find the matched pixel pairs, whose features are then fused. Continuous fusion models outperform the discrete fusion model in all metrics. For BEV detection, the discrete fusion model even has similar scores as the BEV model. This result confirms that fusing image and LIDAR features is not a trivial task.

When geometric feature is removed from MLP input, the performance of the continuous fusion model significantly drops. However, even when offsets are absent, the continuous fusion model still outperforms the discrete one, which justifies the importance of interpolation by KNN pooling.

Continuous fusion layer has two hyper-parameters, the maximum neighbor distance $d$ and number of nearest neighbors $k$. Setting a threshold on the distance to selected neighbors prevents propagation of wrong information from far away neighbors. However, as shown in Table \ref{tab:ablation}, the model is insensitive to such threshold (k=1, d=+inf). One reason might be that the model learns to ``ignore" neighbors when their distance is too far. When the number of nearest neighbor is increased from 1 to 3, the performance is even worse. A possible reason is that larger $k$ will lead to more distant neighbors, which have less prediction power than close neighbors. Empirically, for any of distance threshold chosen, the model with KNN pooling consistently outperforms the model without KNN pooling.

\subsection{TOR4D BEV Object Detection}

%=================== kitti-evaluation-results ================
\begin{table*}[t]
	\begin{center}
		\begin{tabular}{|c|cc|cc|cc|}
			\hline
			\multirow{2}{*}{Model} & \multicolumn{2}{|c|}{Vehicle} & \multicolumn{2}{|c|}{Pedestrian} & \multicolumn{2}{|c|}{Bicyclist}\\
			\cline{2-7}
			& AP$_{0.5}$ & AP$_{0.7}$ & AP$_{0.3}$ & AP$_{0.5}$ & AP$_{0.3}$ & AP$_{0.5}$\\
			\hline
			PIXOR & 91.35 & 79.37 & n/a & n/a & n/a & n/a \\
			Ours (BEV only) & 93.26 & 81.41 & 78.87 & 72.46 & 70.97 & 57.63 \\
			Ours (Continuous Fusion) & \bf 94.94 & \bf 83.89 & \bf 82.32 & \bf 75.34 & \bf 74.08 & \bf 59.83 \\
			\hline
		\end{tabular}
		\caption{Evaluation of multi-class BEV object detection on TOR4D dataset. We compare the continuous fusion model with the BEV baseline, and a recent LIDAR based detector PIXOR \cite{pixor}. The evaluation is conducted on the front view, with 100 meters range along the $x$ axis and 40 meters range along the $y$ axis in LIDAR space.}
		\label{tab:tor4d}
	\end{center}
\end{table*}
%==============================================================

\subsubsection{Dataset and metrics.}
\begin{figure}[t]
	\centering
	\begin{subfigure}{0.32\textwidth}
		\includegraphics[width=\textwidth]{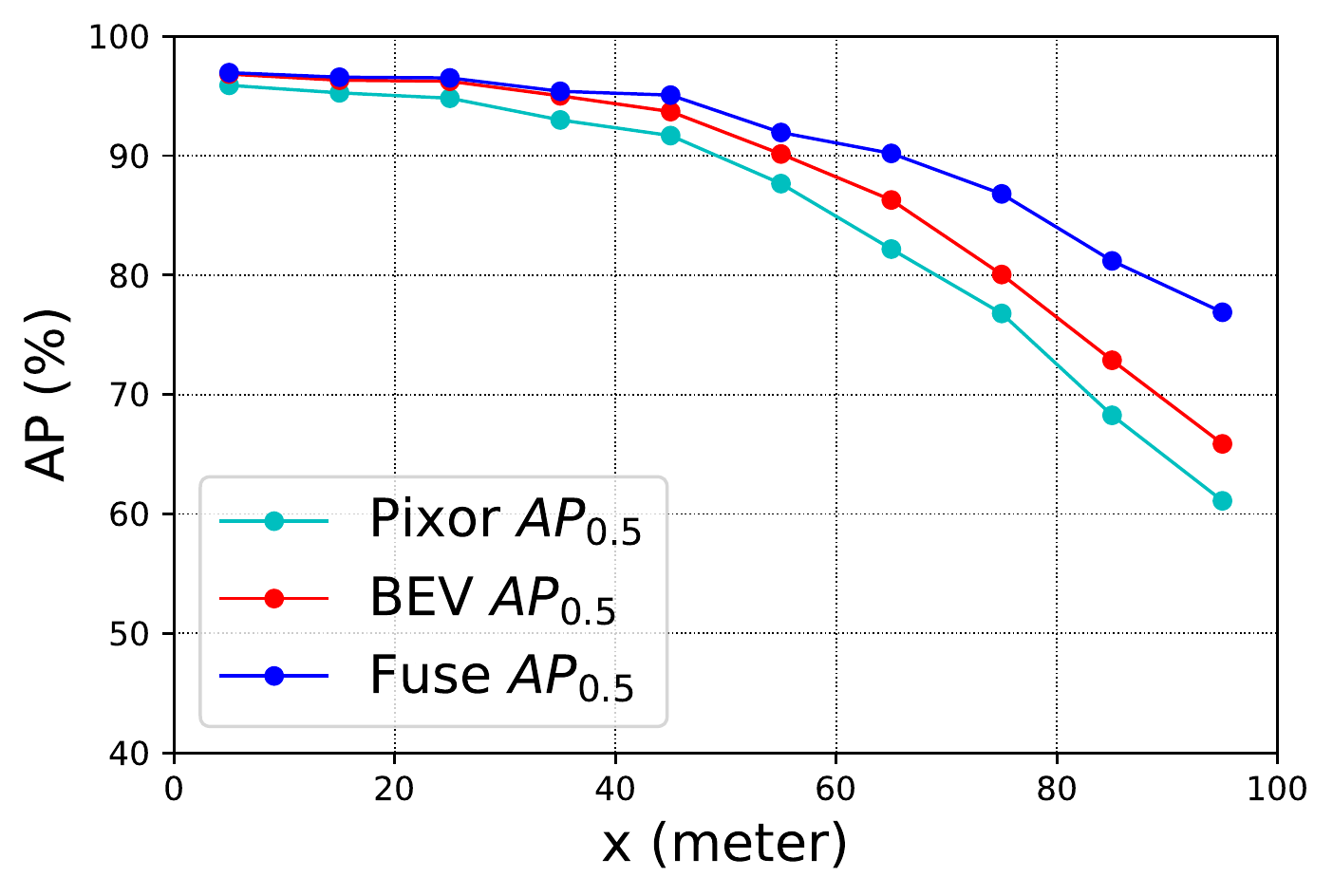}
		\caption{Vehicle AP$_{0.5}$}
		\label{fig:car_ap05_distance}
	\end{subfigure}
	\begin{subfigure}{0.32\textwidth}
		\includegraphics[width=\textwidth]{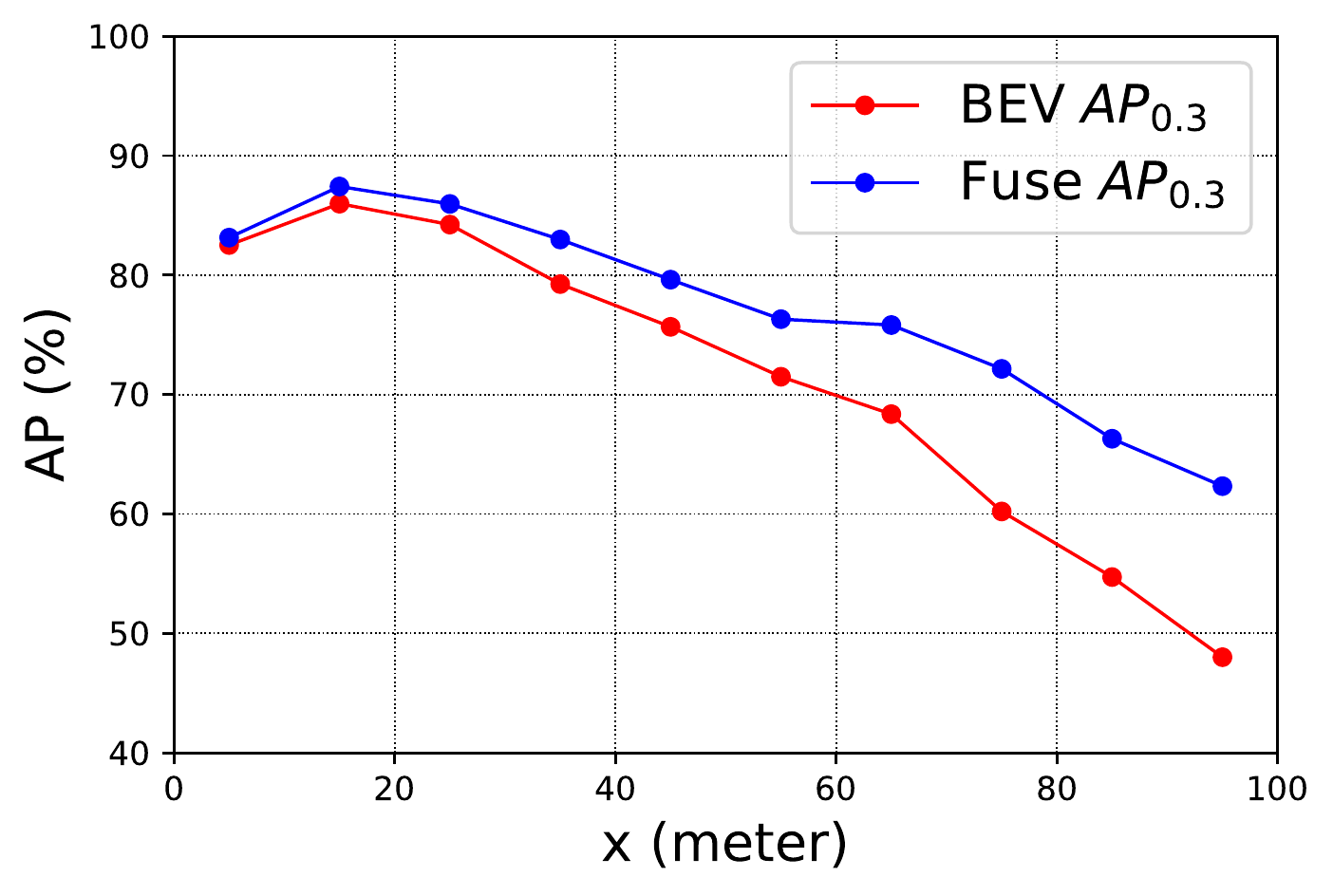}
		\caption{Pedestrian AP$_{0.3}$}
		\label{fig:ped_ap03_distance}
	\end{subfigure}
	\begin{subfigure}{0.32\textwidth}
		\includegraphics[width=\textwidth]{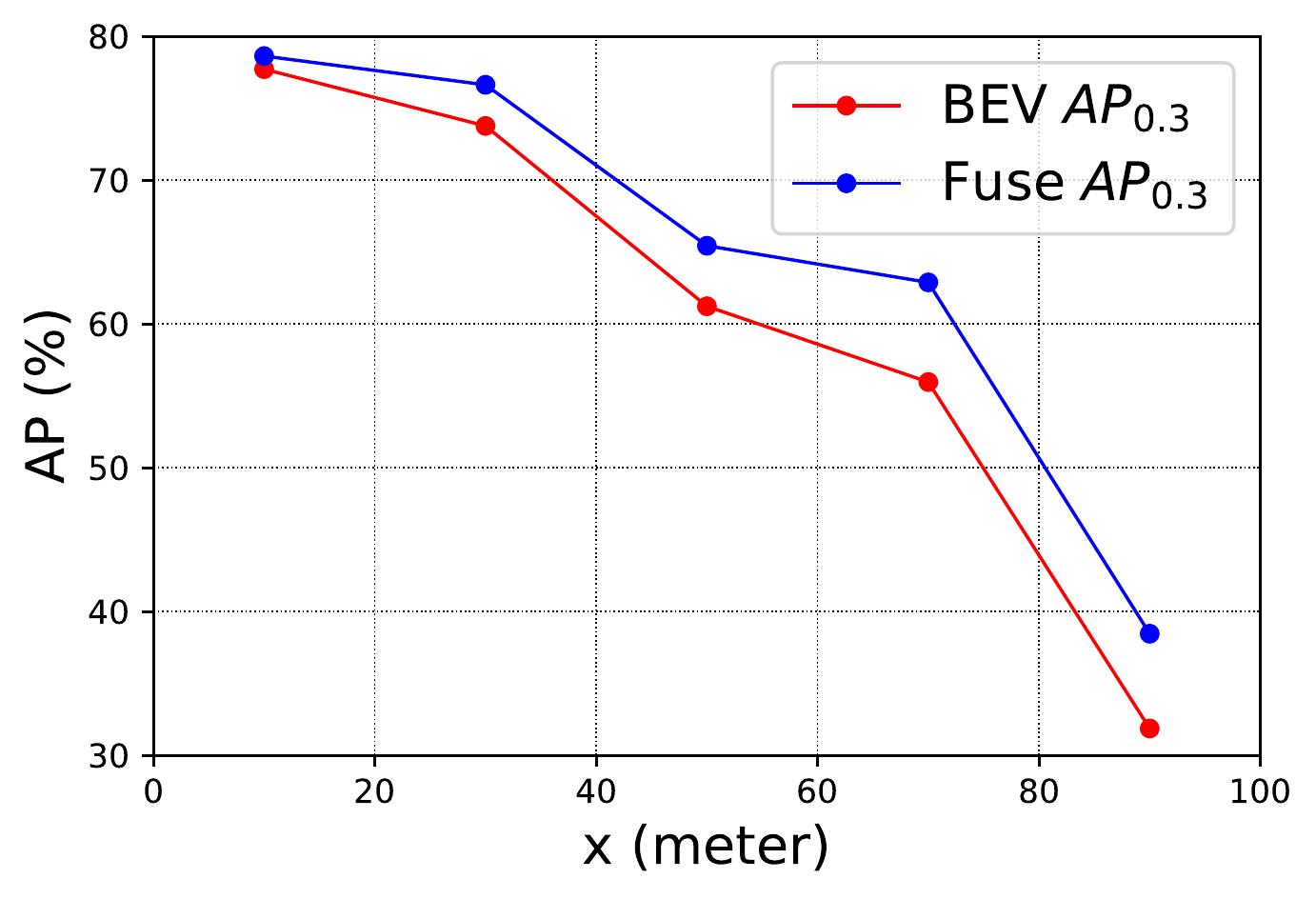}
		\caption{Bicyclist AP$_{0.3}$}
		\label{fig:bike_ap03_distance}
	\end{subfigure}
	\begin{subfigure}{0.32\textwidth}
		\includegraphics[width=\textwidth]{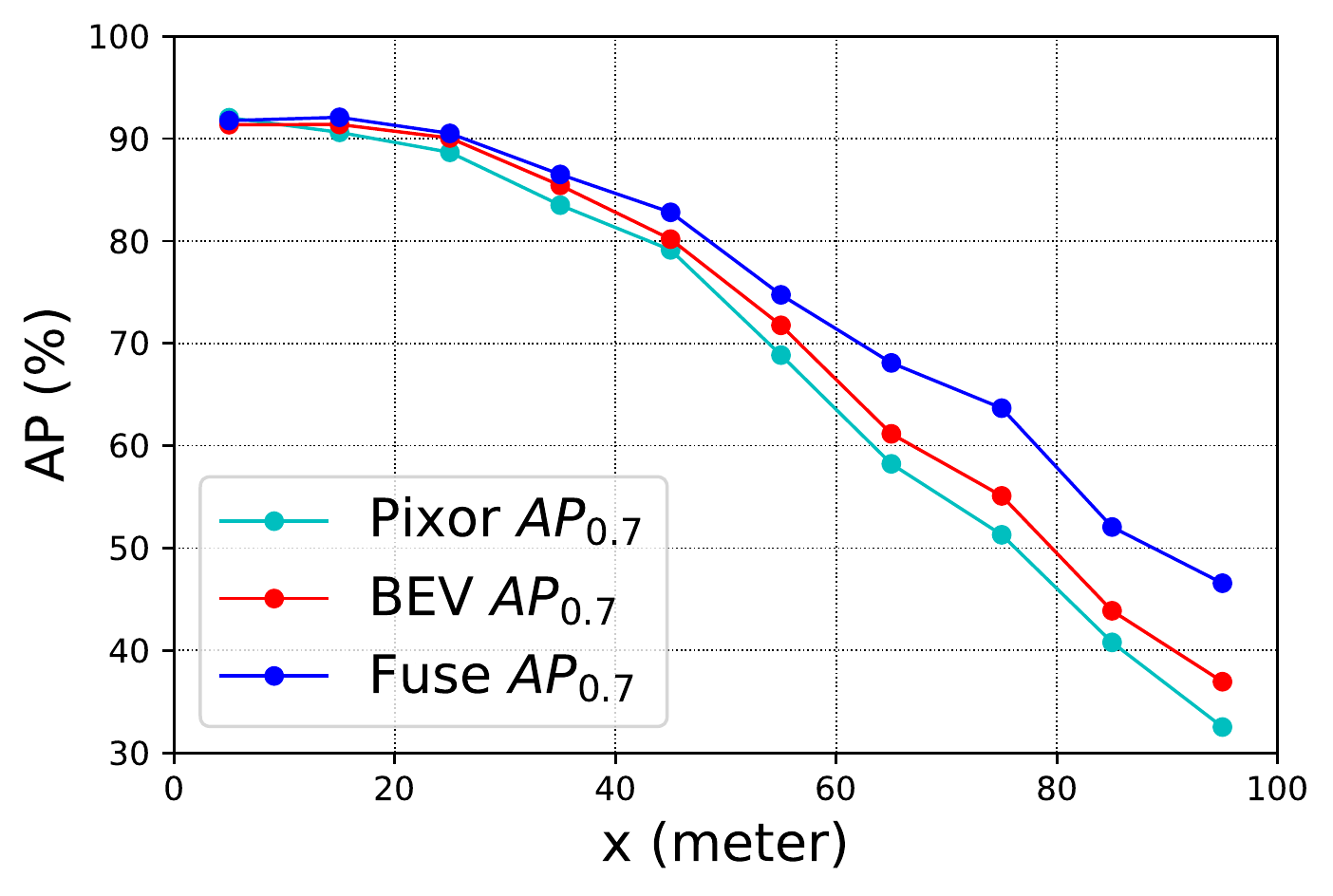}
		\caption{Vehicle AP$_{0.7}$}
		\label{fig:car_ap07_distance}
	\end{subfigure}
	\begin{subfigure}{0.32\textwidth}
		\includegraphics[width=\textwidth]{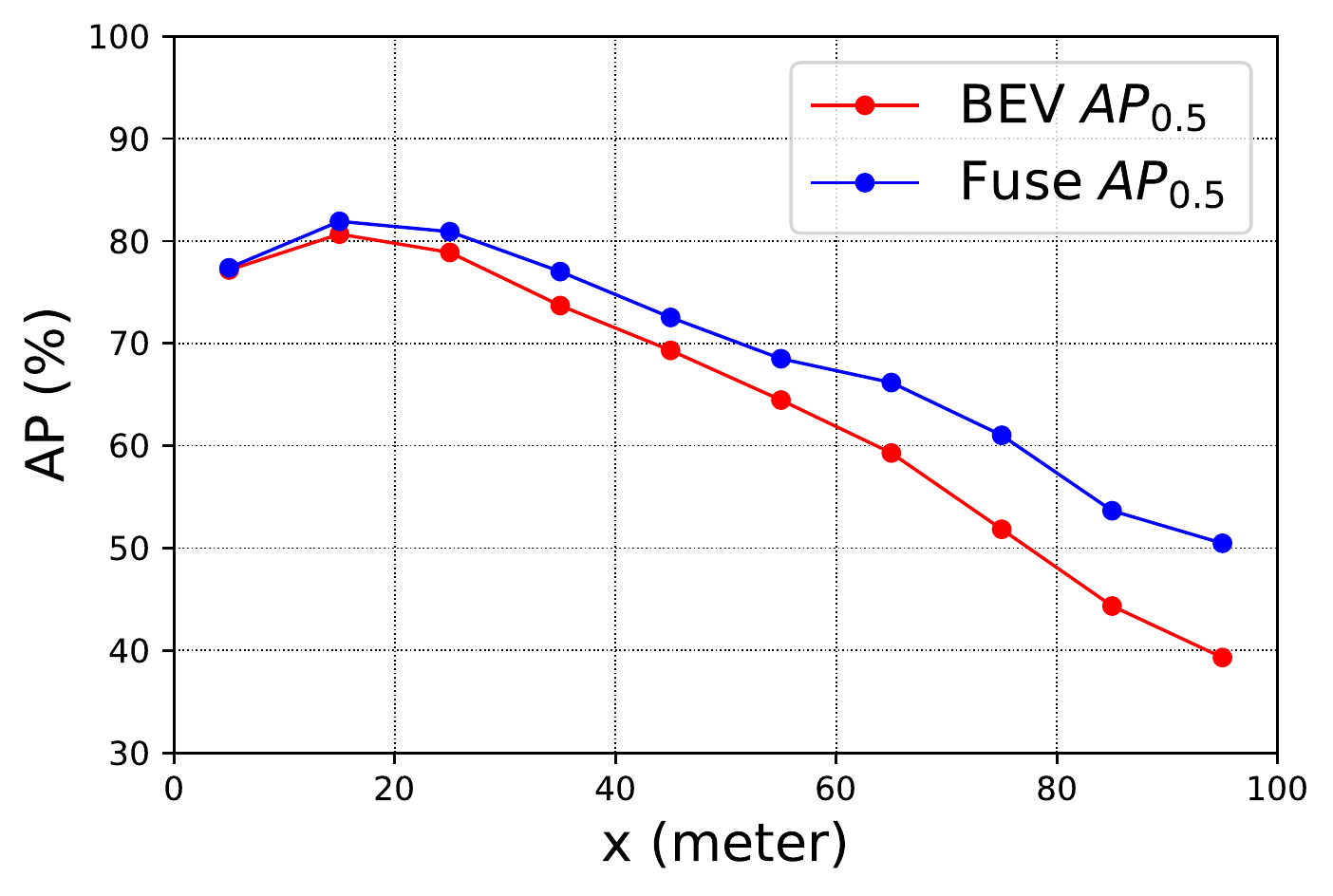}
		\caption{Pedestrian AP$_{0.5}$}
		\label{fig:ped_ap05_distance}
	\end{subfigure}
	\begin{subfigure}{0.32\textwidth}
		\includegraphics[width=\textwidth]{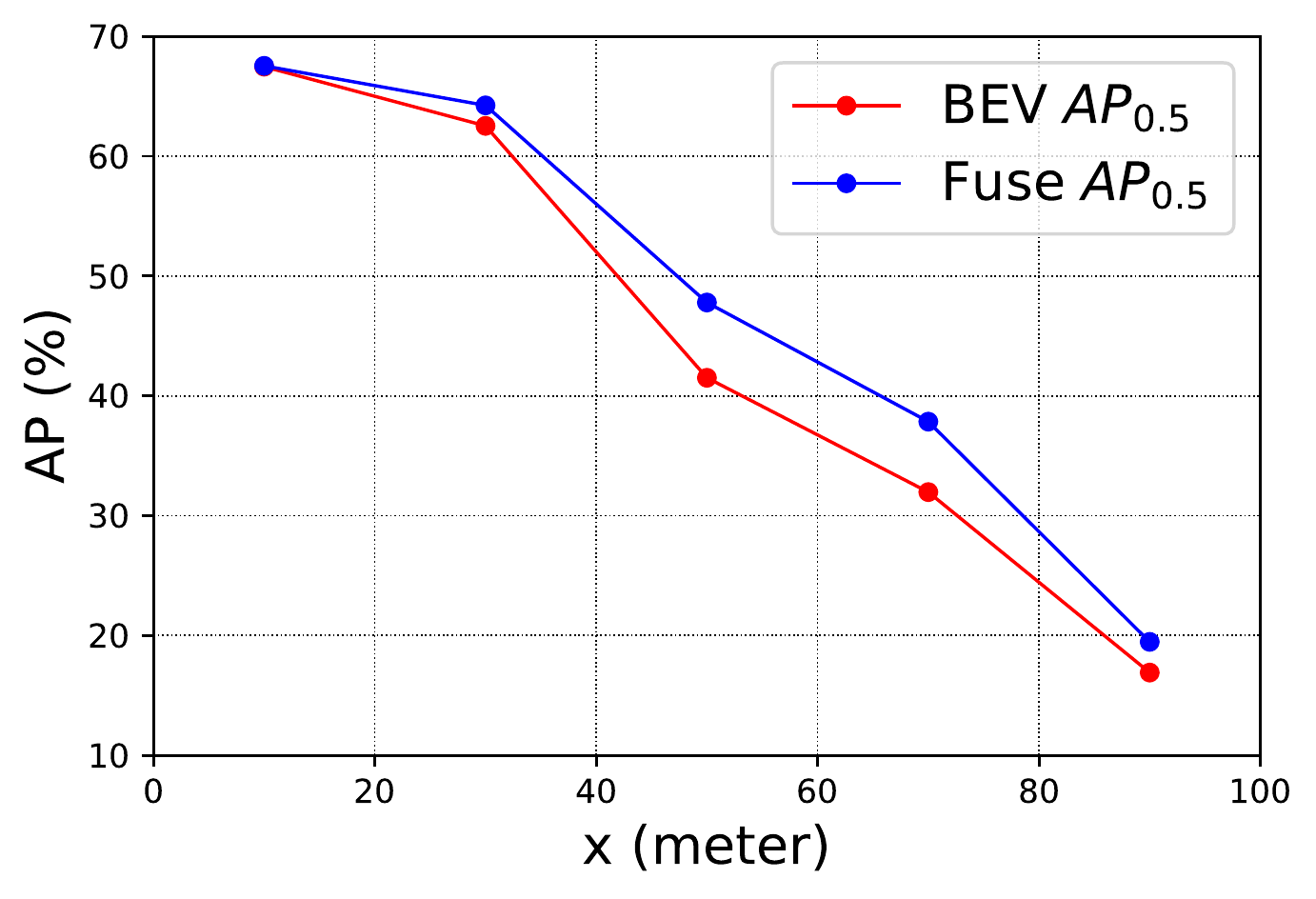}
		\caption{Bicyclist AP$_{0.5}$}
		\label{fig:bike_ap05_distance}
	\end{subfigure}
\caption{Piecewise AP of multi-class BEV object detection on TOR4D dataset. For vehicle and pedestrian, each point is computed within a 10-meter range along the $x$ axis in LIDAR space. For bicyclist, each point is computed within a 20-meter range because there are fewer targets. Our continuous fusion model, its BEV baseline, and PIXOR \cite{pixor} are compared. When $x$ is very small, fusion model and LIDAR models have similar performance. The advantage of fusion model increases when $x$ increases.}
\label{fig:ap_distance}
\end{figure}

We evaluate our model on TOR4D, a newly collected large scale 3D object detection dataset with long range object labels annotated. The training set has more than 1.2 million frames, extracted from $5$K sequences. The validation set contains 5969 frames, extracted from 500 sequences. The sampling rate for training and validation frames are 10 Hz and 0.5 Hz, respectively. The data is collected on roads in North-American cities. Both LIDAR and images are collected, and BEV bounding box annotations are provided over 100 meters.

The model architecture and training procedure are similar to those on KITTI. One major difference is the input size. On TOR4D our BEV input spans 100 meters in front direction. To compensate for the extra time cost caused by larger input, we reduce the feature dimension of the BEV network. The input image size is $1200\times 1920$, which is also larger than KITTI images. To achieve real-time efficiency, we only use a narrow image crop of size $224\times 1920$. Overall, after these changes our TOR4D model is even faster than the KITTI model, running at $0.05$ second per frame.

A multi-class BEV object detection model is trained on the dataset. The model detects three classes, including vehicle, pedestrian and bicyclist. We changed the detection header to have multiple classification and regression outputs, one for each class. The two $z$ axis related regression terms are removed from the loss function. 2D rotated IoU on BEV is used as the evaluation metric. AP$_{0.5}$ and AP$_{0.7}$ are used for vehicle class, and AP$_{0.3}$ and AP$_{0.5}$ are used for the other two classes. On this large scale dataset we do not find significant benefit from regularization techniques, such as data augmentation and dropout. We thus do not use these techniques. The model is trained with Adam optimizer \cite{adam} at 0.001 learning rate for 1 epoch, and then 0.0001 learning rate for another 0.4 epoch.

\subsubsection{Evaluation results}
We compare the continuous fusion model with two baseline models. One is a BEV model which is basically the BEV stream of the fusion model. The other one is a recent state-of-the-art BEV detection model, PIXOR \cite{pixor}, which is based on LIDAR input only. We evaluate the models over the range of $0\le x\le100$ and $-40\le y\le40$, where $x$ and $y$ are axes in the LIDAR space.

Our continuous fusion model significantly outperforms the other two LIDAR based methods on all classes(Table 3). To better illustrate the performance of our model on long range object detection, we compute range based piecewise AP for the three classes (Figure \ref{fig:ap_distance}).
Each point is computed over 10-meter range for vehicle and pedestrian and 20-meter range for bicyclist along the $x$ axis. For all classes and metrics, the continuous fusion model outperforms BEV and PIXOR \cite{pixor} at most ranges. The advantage of fusion models generally increases when $x$ increases.

\subsection{Qualitative Results and Discussion}
\begin{figure*}[t]
	%\footnotesize
	\setlength\tabcolsep{0.5pt} % default value: 6pt
	\renewcommand{\arraystretch}{0.8}
	\begin{tabular}{ccc}
		\includegraphics[width=.33\linewidth]{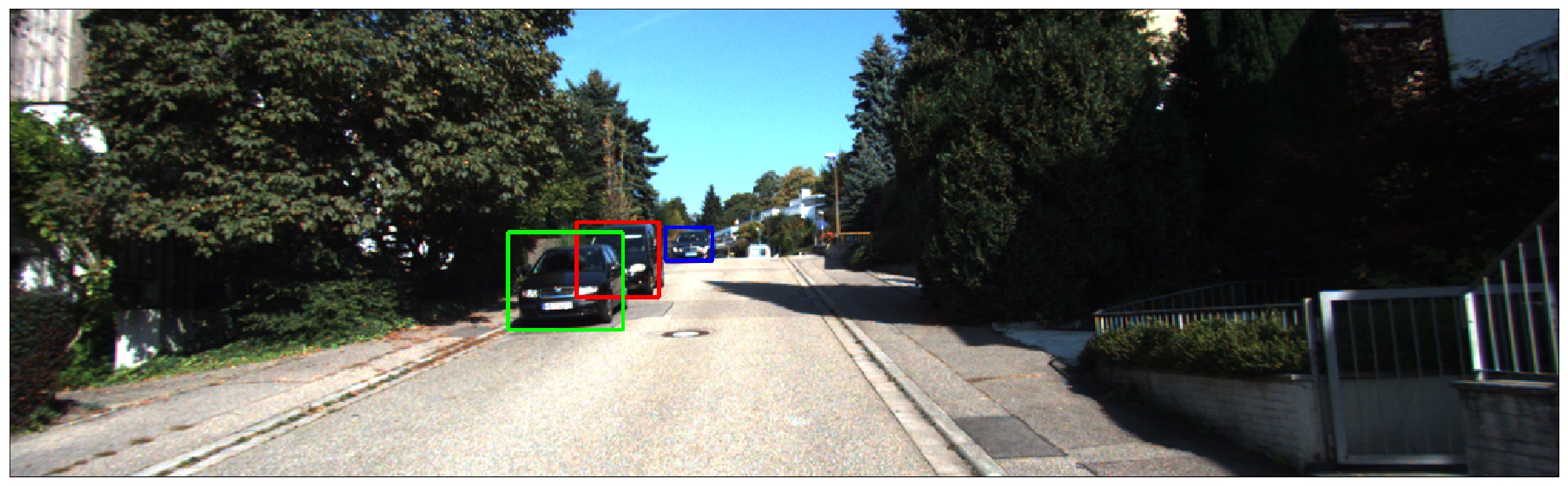} & 
		\includegraphics[width=.33\linewidth]{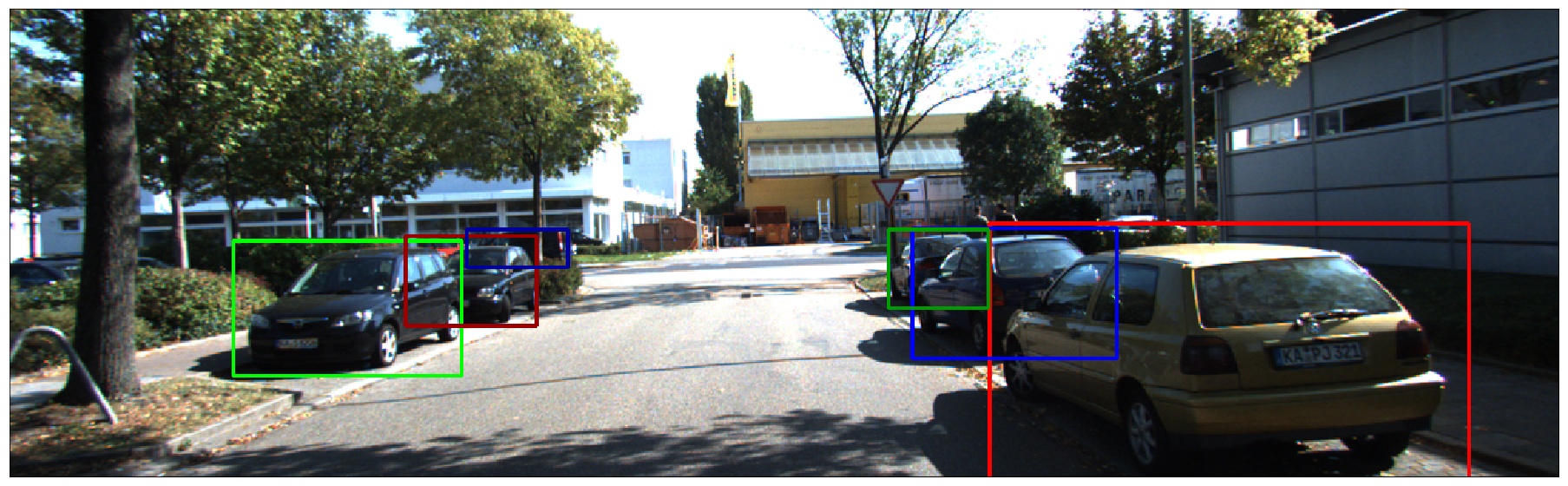} & 
		\includegraphics[width=.33\linewidth]{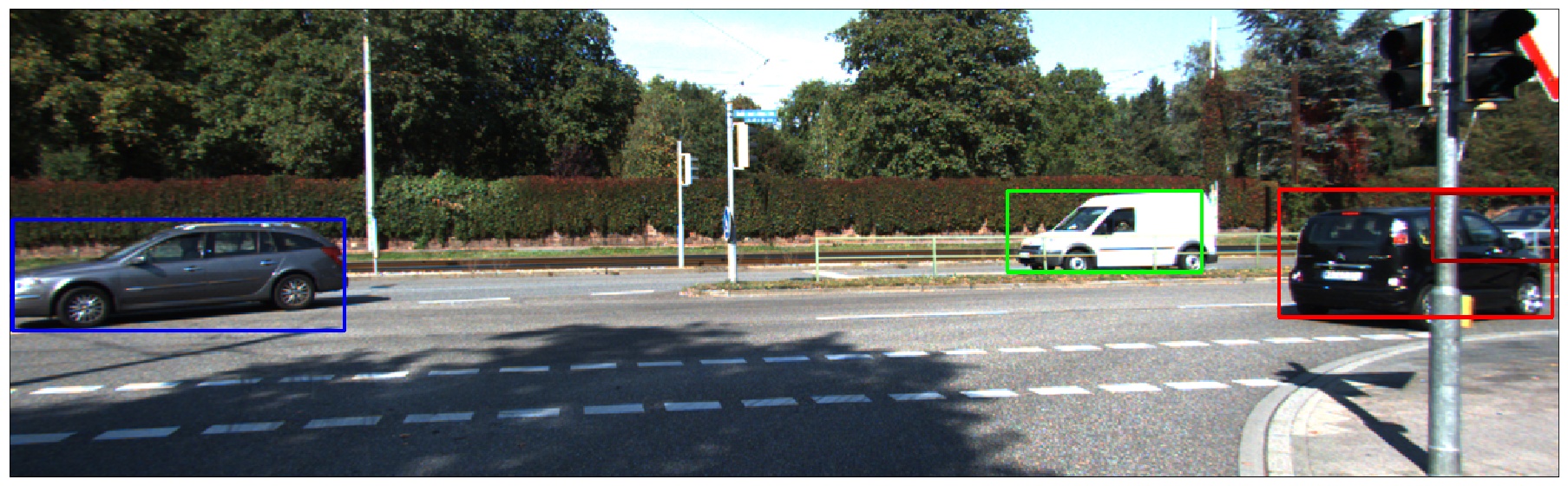} \\
		\adjincludegraphics[width=.33\linewidth, trim={{.1\width} {.3\height} {.1\width} {.35\height}}, clip]{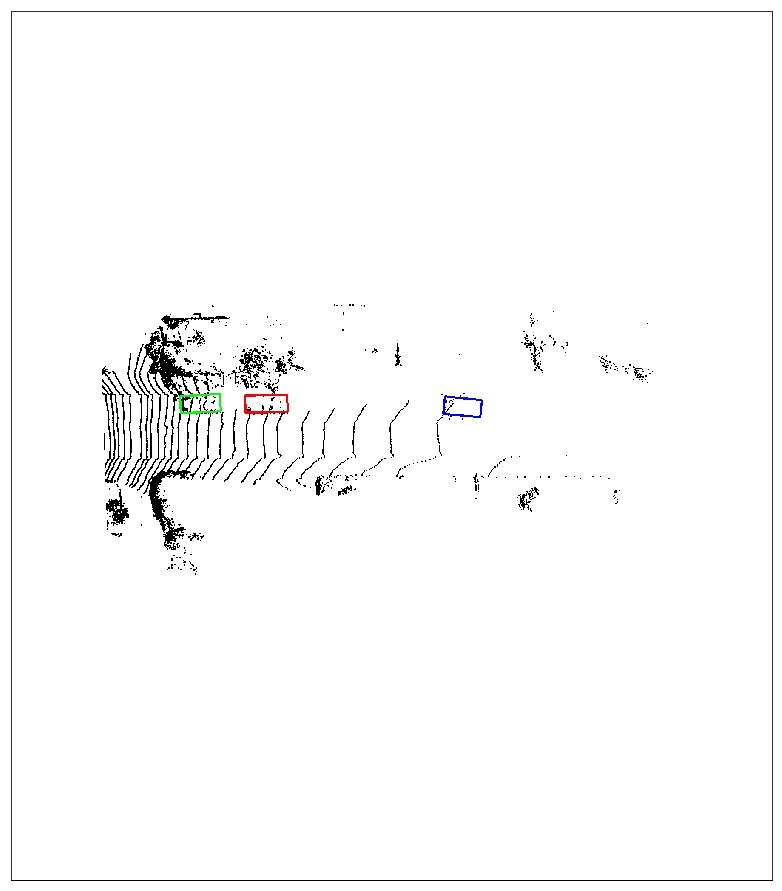} & 
		\adjincludegraphics[width=.33\linewidth, trim={{.1\width} {.3\height} {.1\width} {.35\height}}, clip]{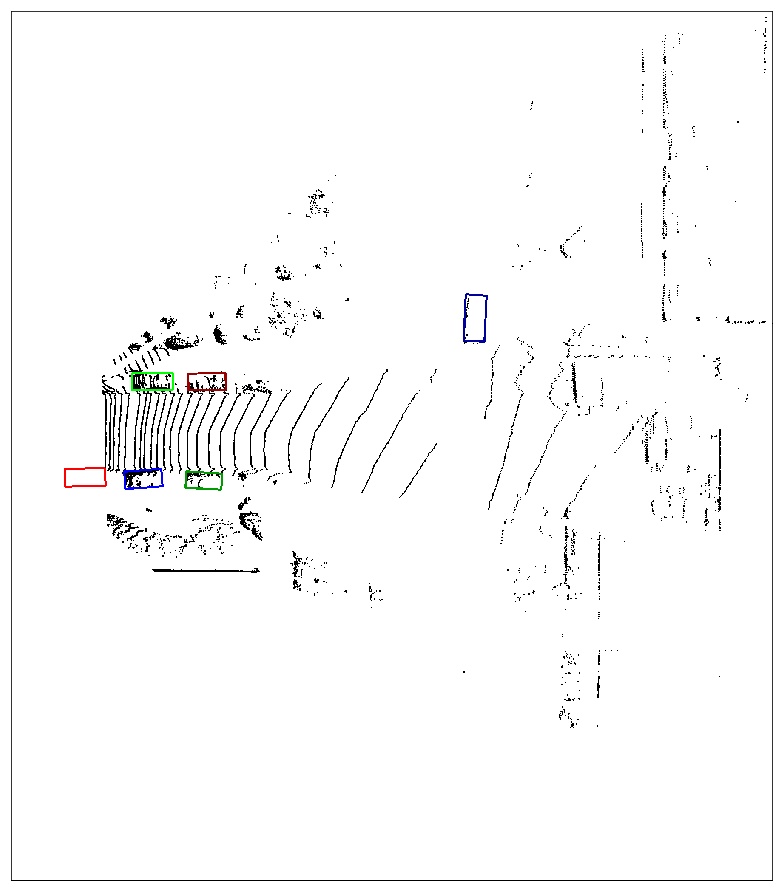} & 
		\adjincludegraphics[width=.33\linewidth, trim={{.1\width} {.3\height} {.1\width} {.35\height}}, clip]{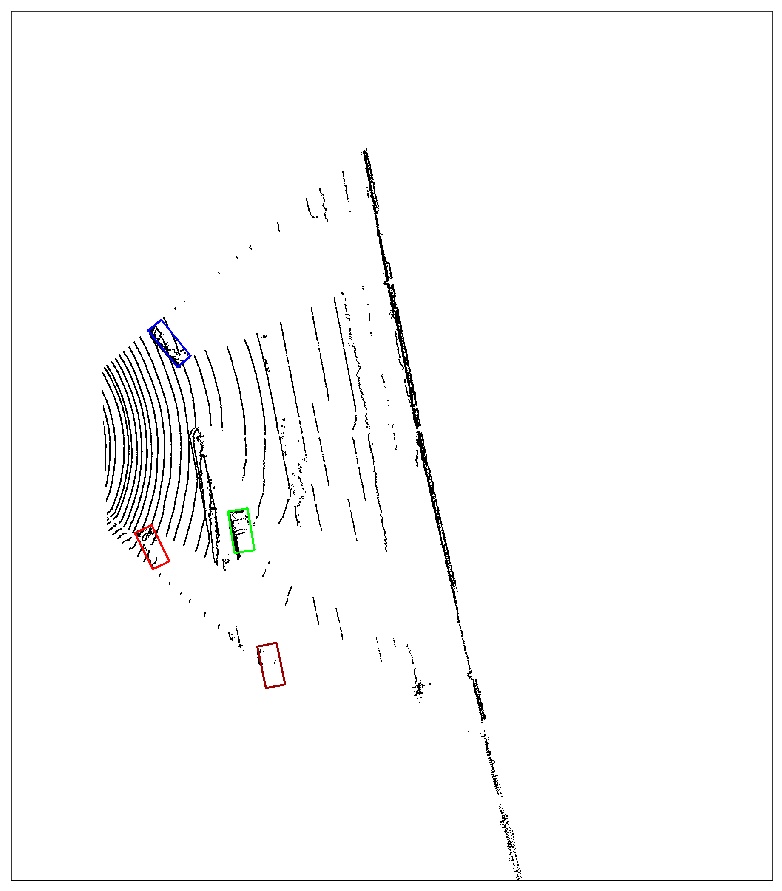} \\  
		\includegraphics[width=.33\linewidth]{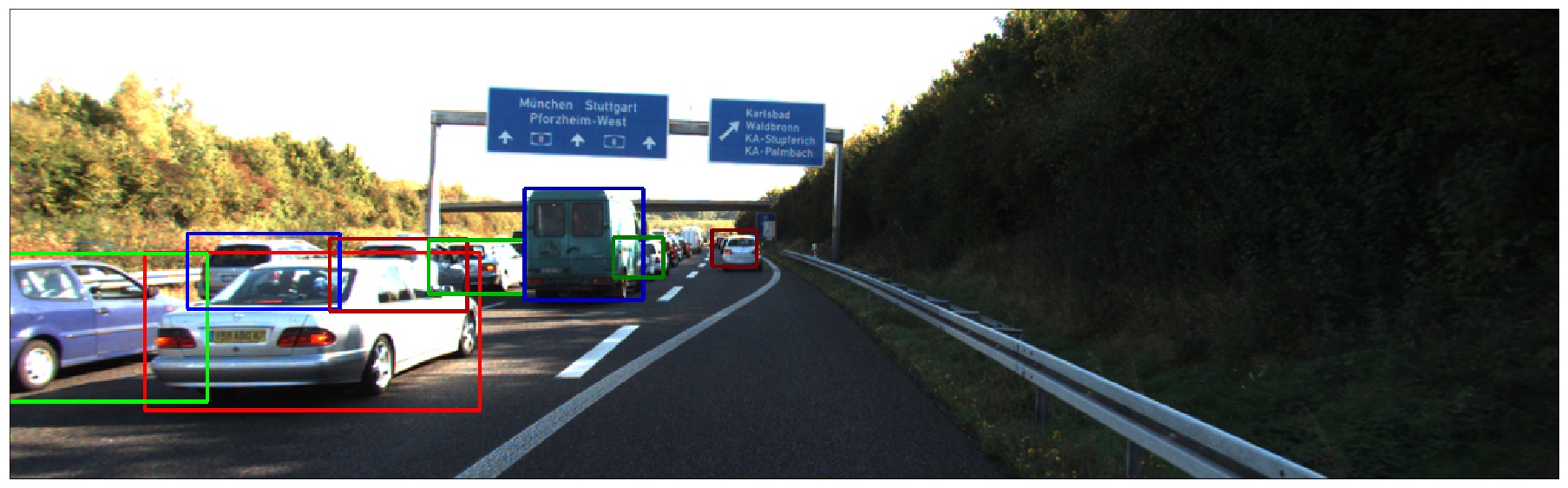} & 
		\includegraphics[width=.33\linewidth]{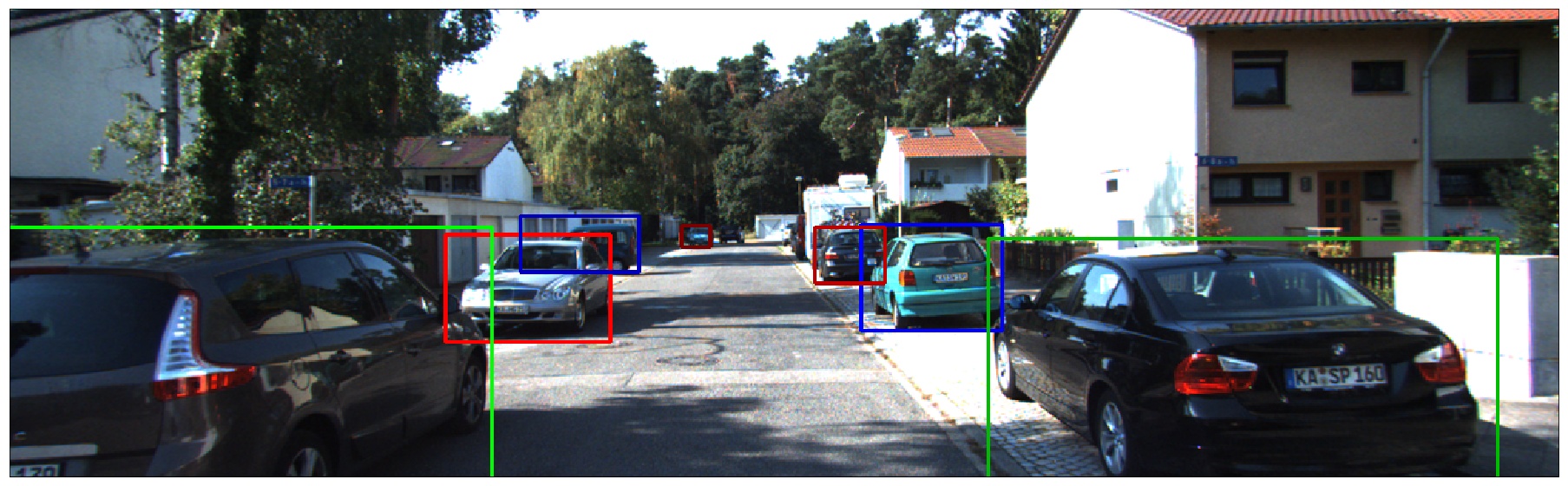} & 
		\includegraphics[width=.33\linewidth]{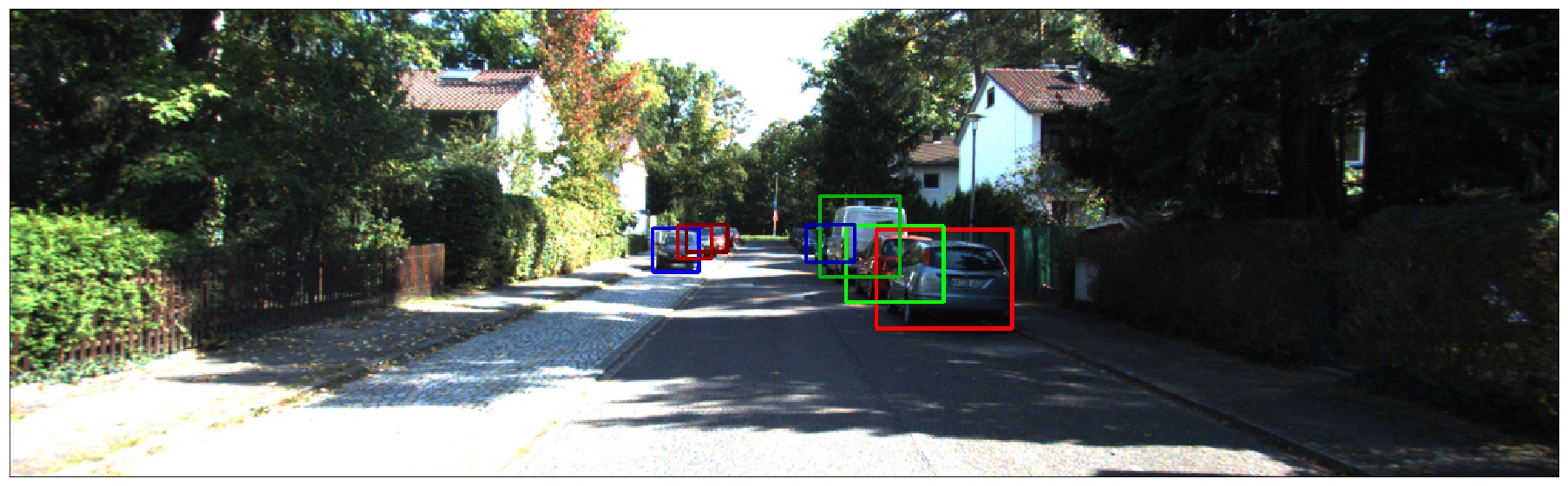} \\
		\adjincludegraphics[width=.33\linewidth, trim={{.1\width} {.3\height} {.1\width} {.35\height}}, clip]{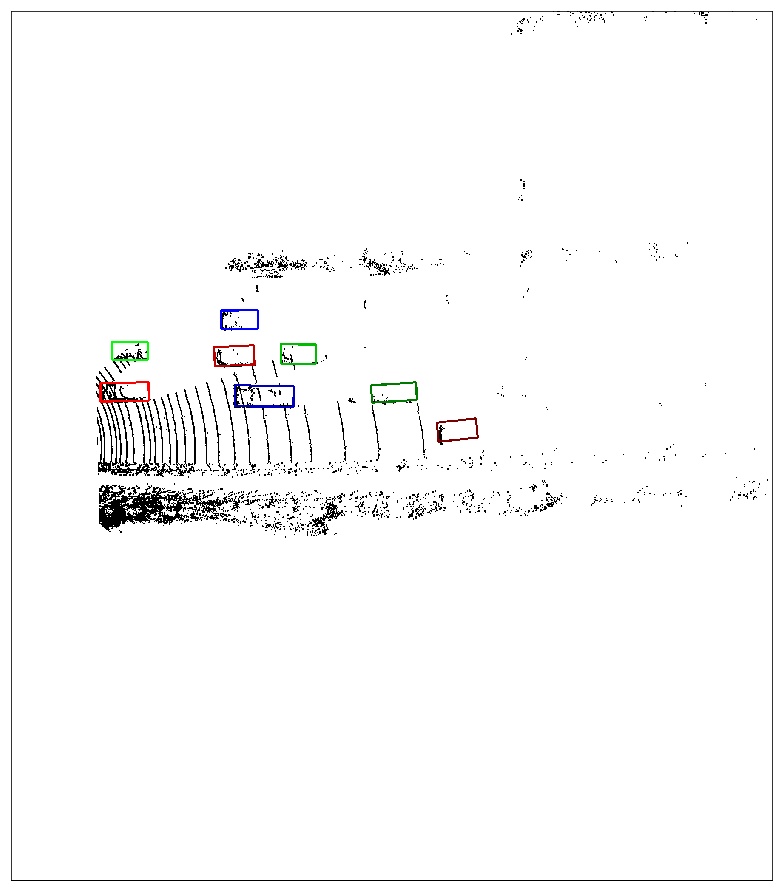} & 
		\adjincludegraphics[width=.33\linewidth, trim={{.1\width} {.3\height} {.1\width} {.35\height}}, clip]{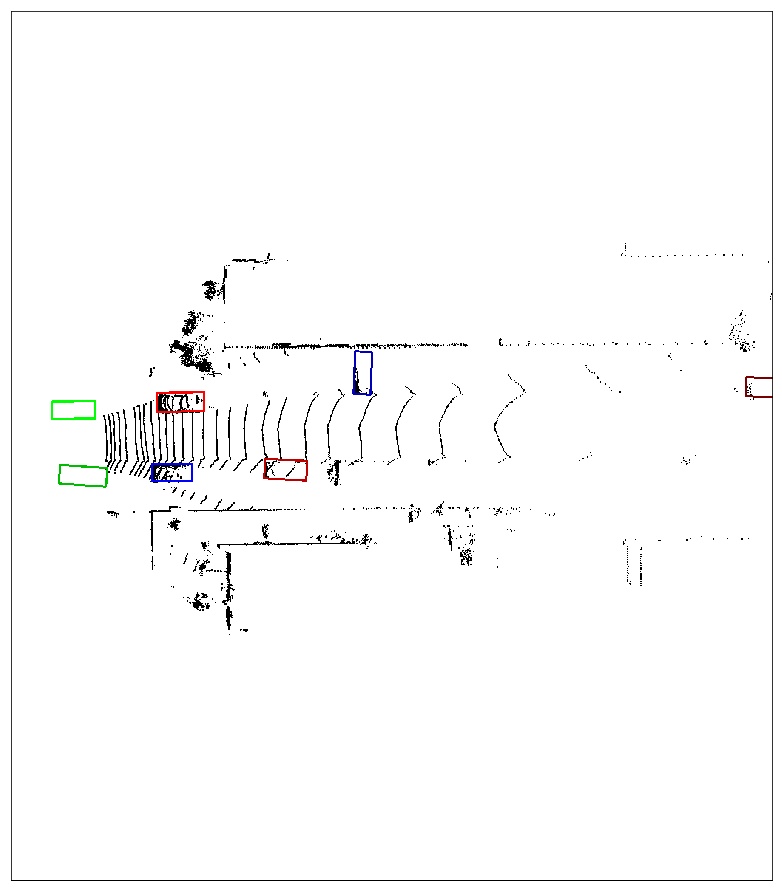} & 
		\adjincludegraphics[width=.33\linewidth, trim={{.1\width} {.3\height} {.1\width} {.35\height}}, clip]{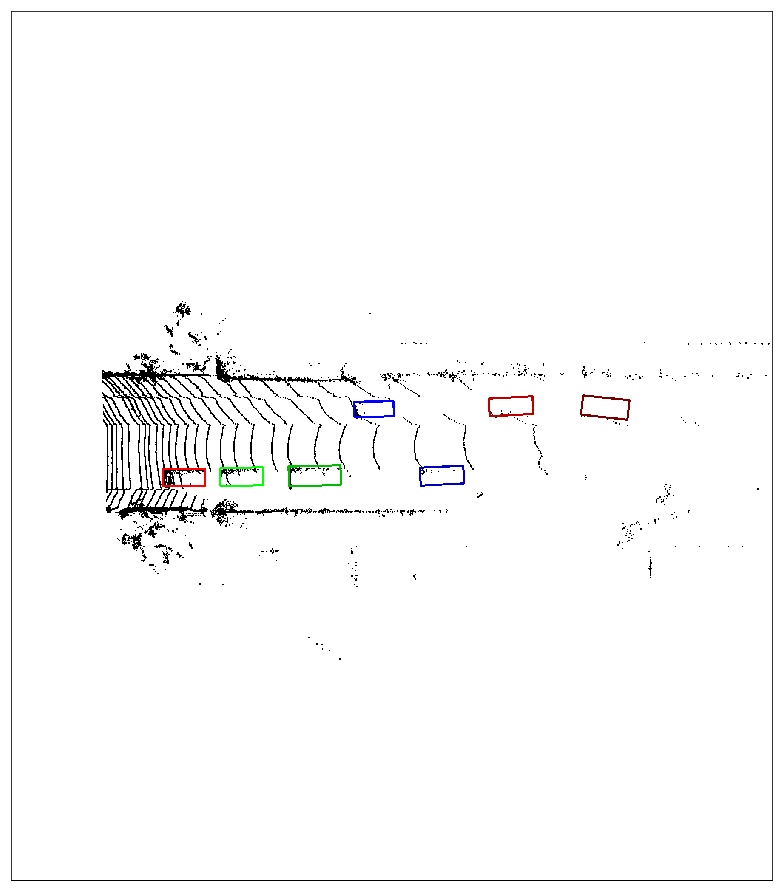} % \\ 
	\end{tabular}
	\caption{Qualitative results on KITTI Dataset. The BEV-image pairs and the detected bounding boxes are shown. The 2D bounding boxes are obtained by projecting the 3D detections onto the image. The bounding box of an object on BEV and images are shown in the same color.}
	\label{fig:qr}
\end{figure*}

\begin{figure*}[t]
	%\footnotesize
	\setlength\tabcolsep{0.5pt} % default value: 6pt
	\renewcommand{\arraystretch}{0.8}
	\begin{tabular}{ccc}
               \includegraphics[width=.49\linewidth]{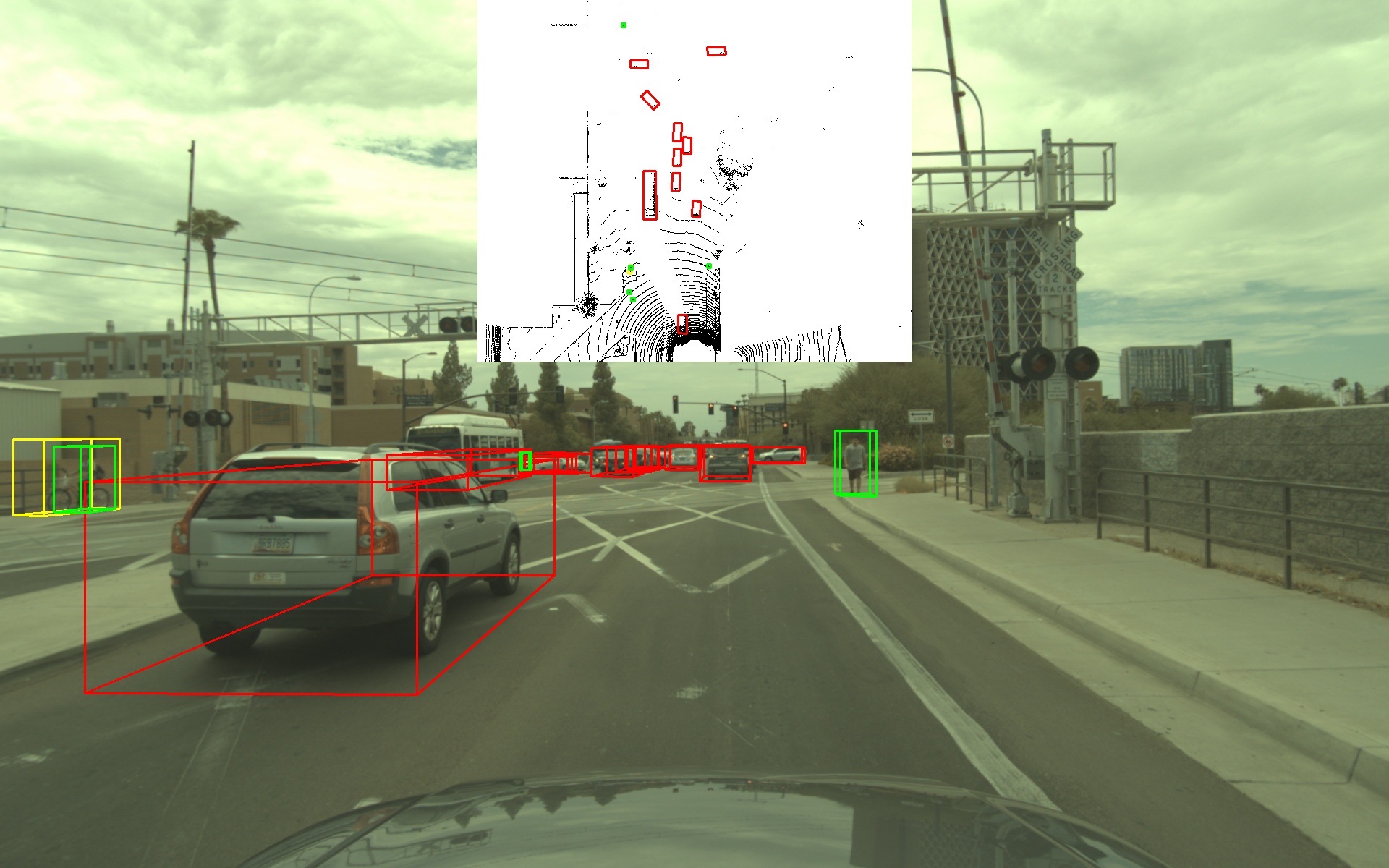} & 
               \includegraphics[width=.49\linewidth]{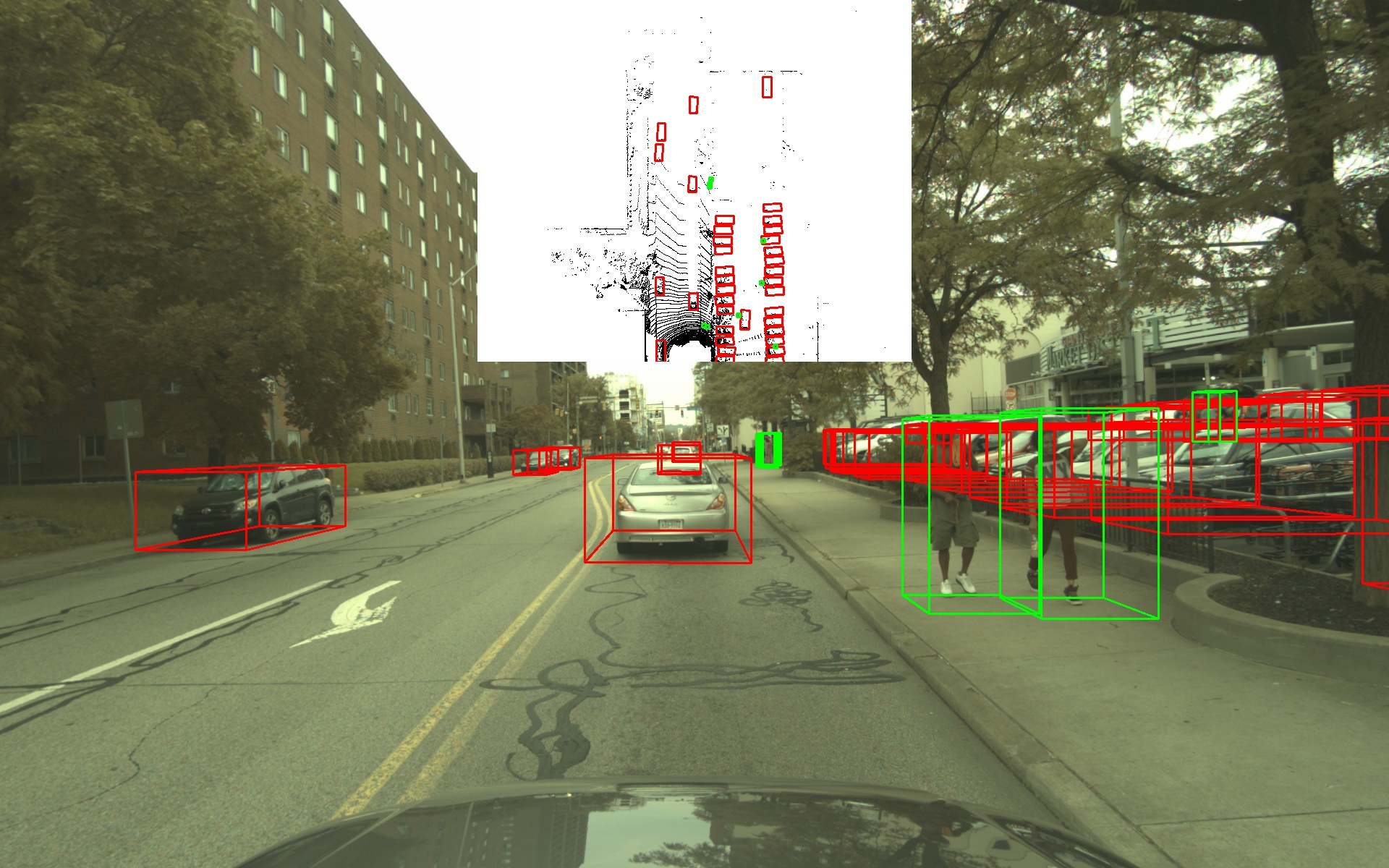} \\
               \includegraphics[width=.49\linewidth]{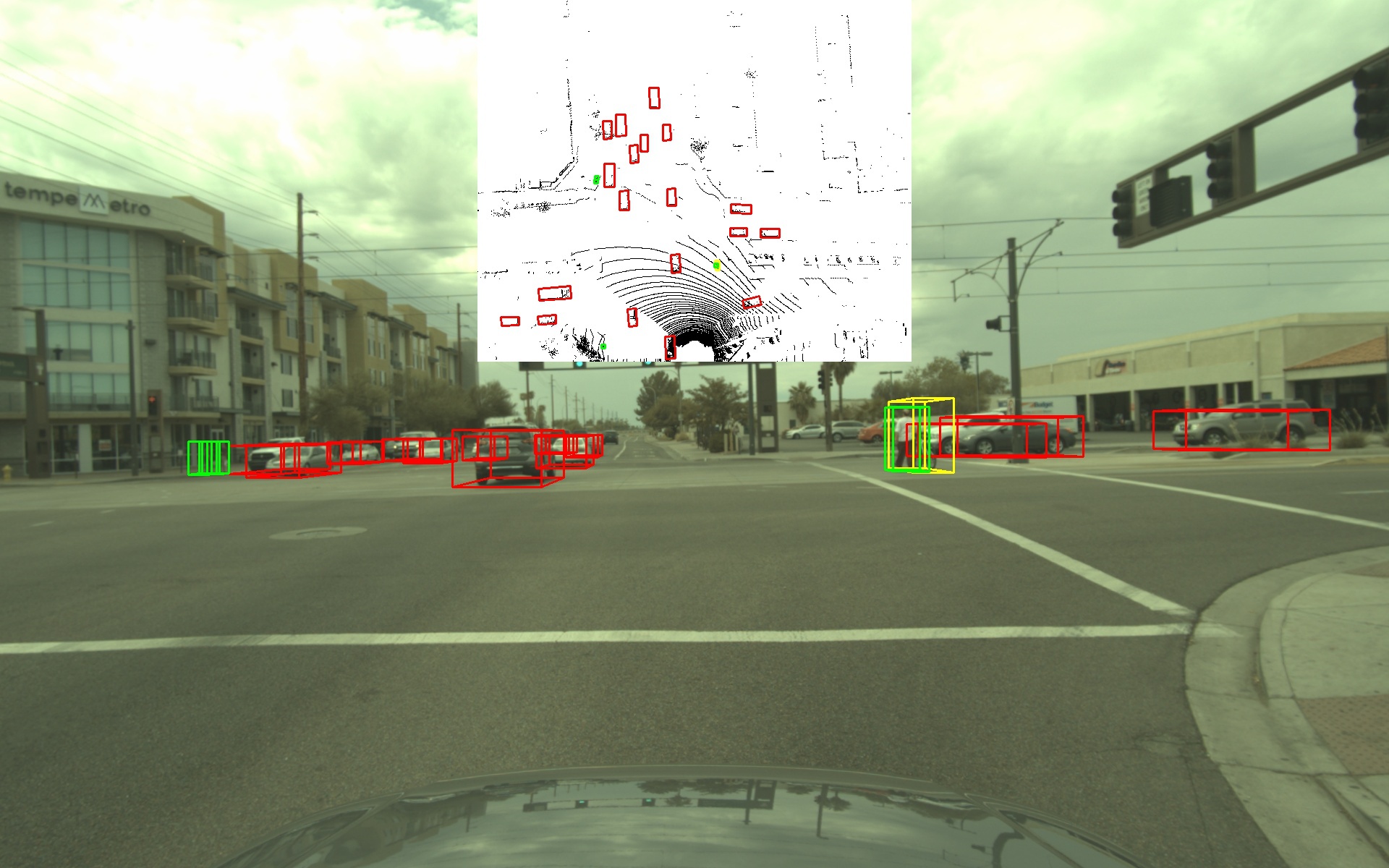} &
               \includegraphics[width=.49\linewidth]{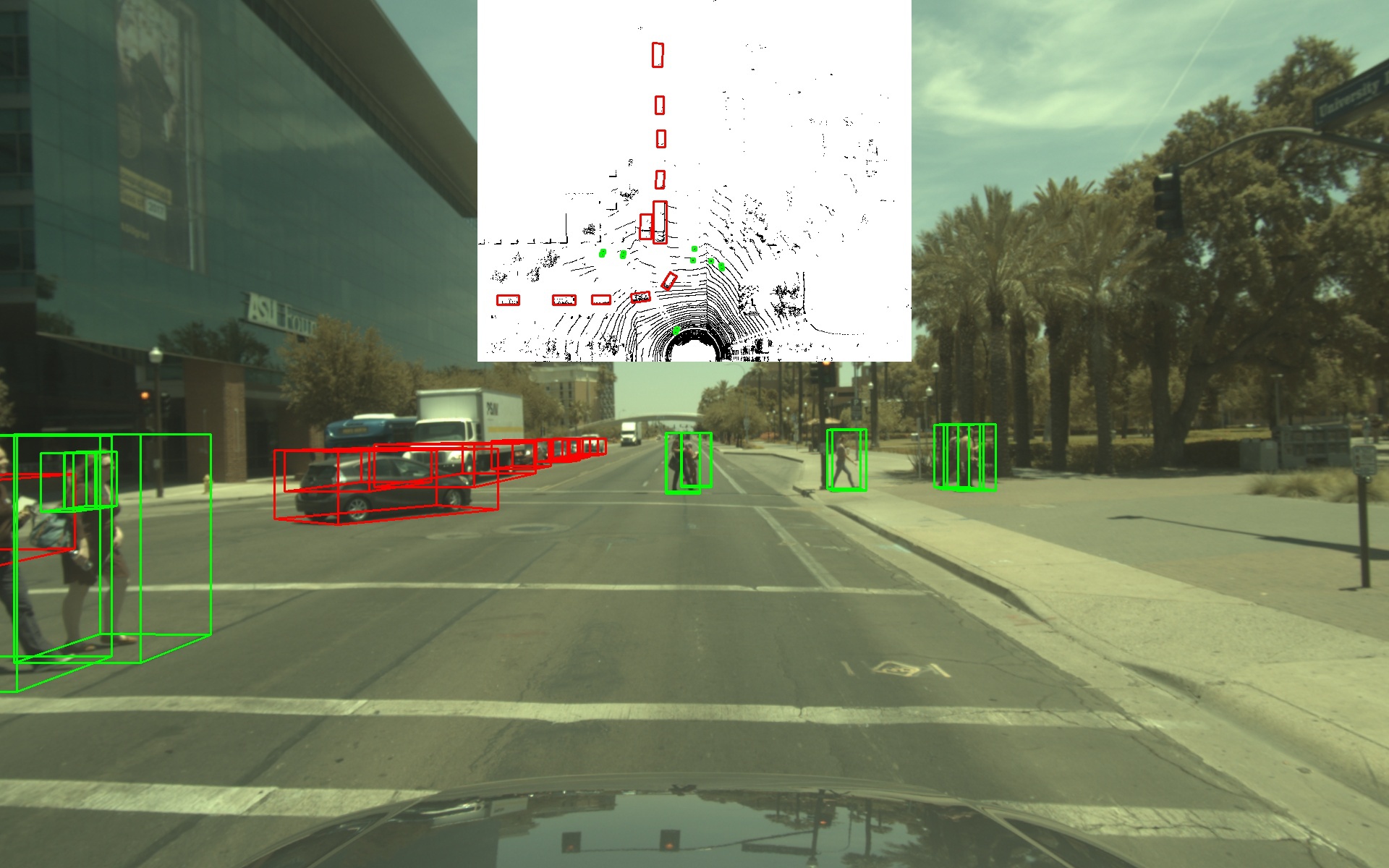} \\ 
               \includegraphics[width=.49\linewidth]{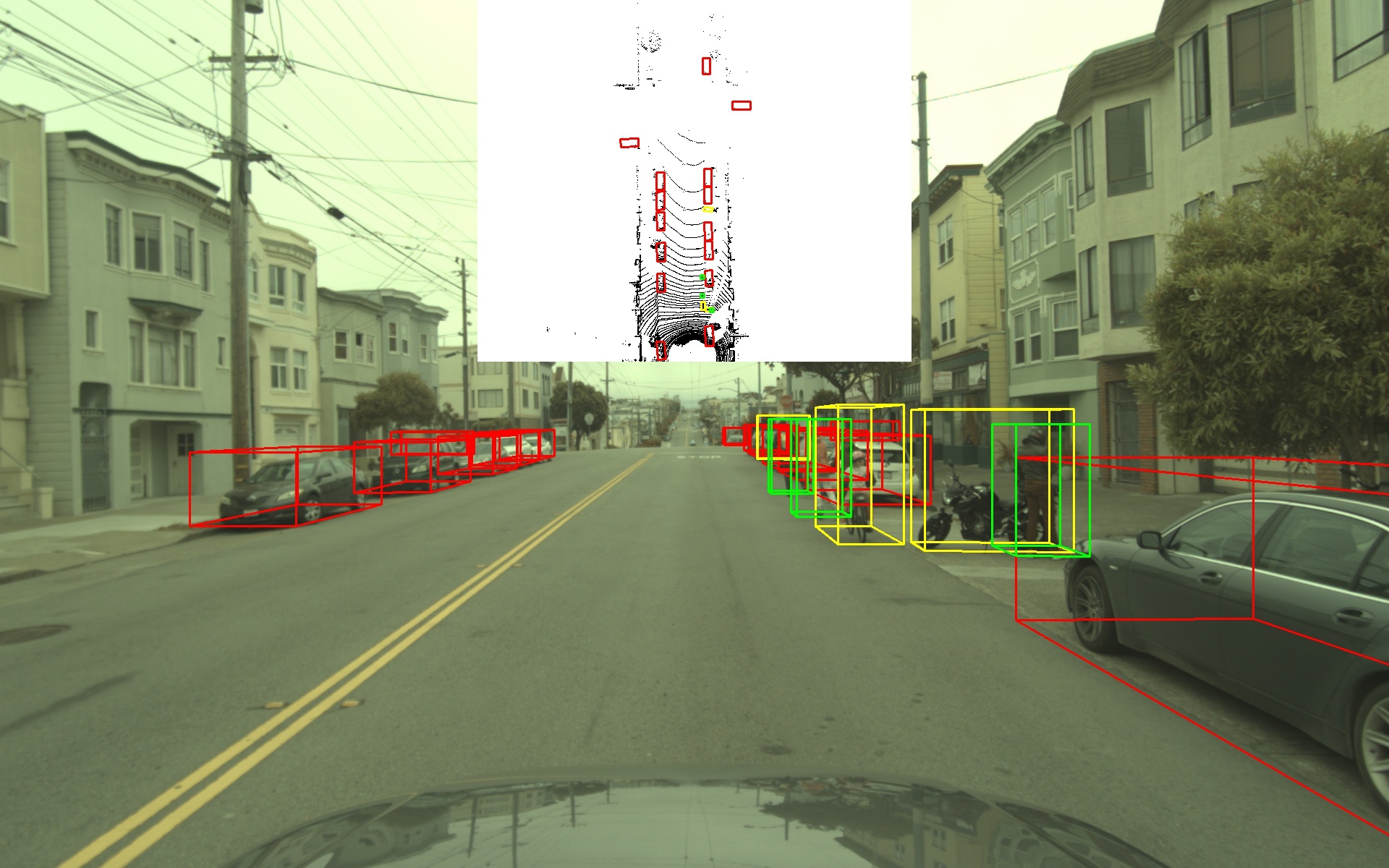} &
               \includegraphics[width=.49\linewidth]{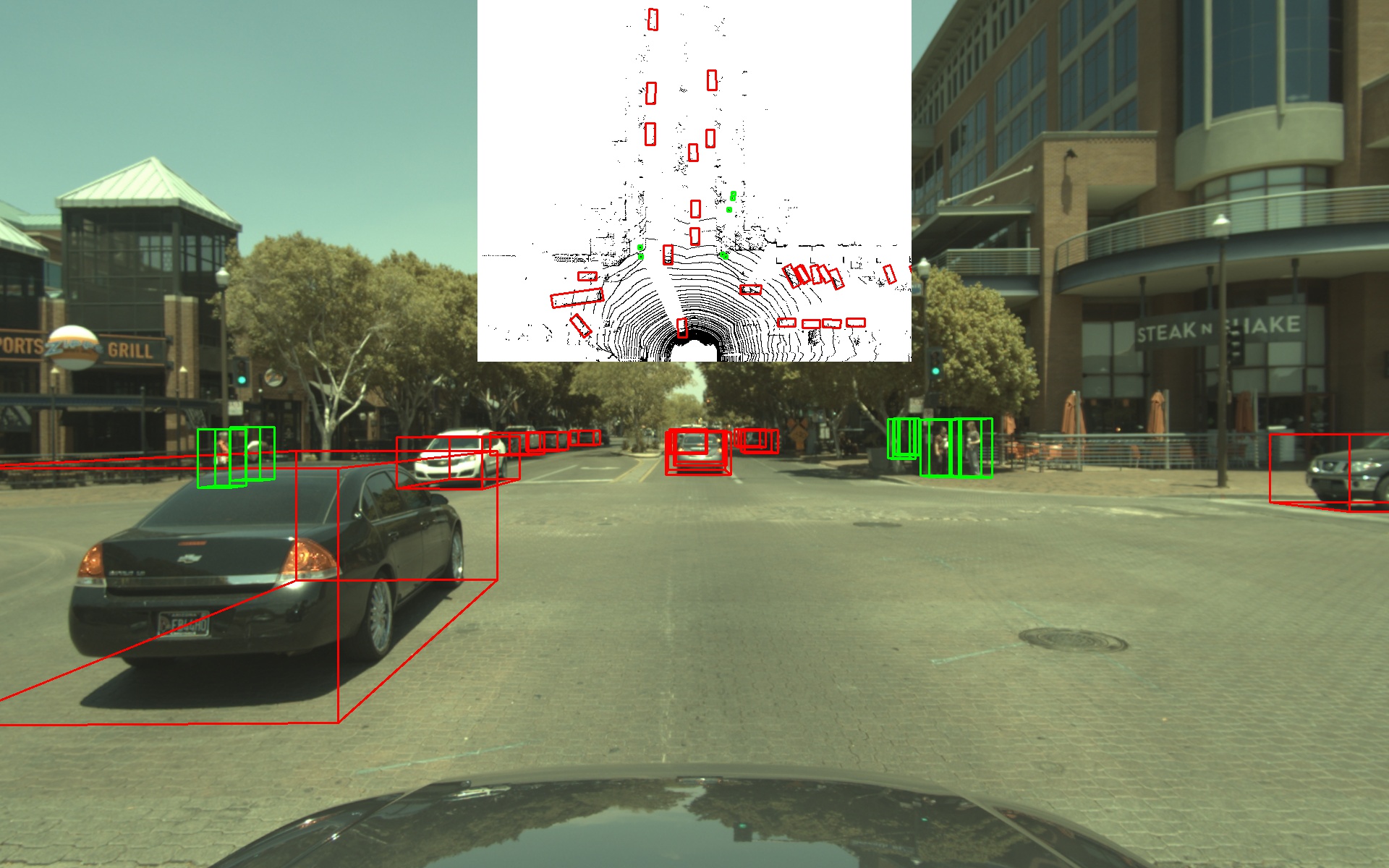} \\
	\end{tabular}
	\caption{Multi-class qualitative results on TOR4D dataset.}
	\label{fig:bev}
\end{figure*}

Qualitative detection results on KITTI are provided in Fig.~\ref{fig:qr}. The BEV and image pairs and detected bounding boxes by our continuous fusion model are shown. The 2D bounding boxes are obtained by projecting 3D detections onto images. The model detects the cars quite well, even when the car is distant or heavily occluded. We also show multi-class qualitative results on the TOR4D dataset in Fig.~\ref{fig:bev}. Because the BEV model does not output height information, we use a fixed $z$ position and height to generate the bounding boxes on images.

Overall, these results demonstrate the excellent scalability of our proposed approach and its superior performance in long range detection. Long range detection is important for autonomous driving. While LIDAR suffers from the extreme data sparsity for distant object detection, high resolution images become a useful information source. High resolution images can be readily incorporated into our model, thanks to the flexibility of continuous fusion layer. Furthermore, no extra image labels are needed because our model can be trained on 3D/BEV detection only.

%!TEX root = top.tex
\section{Conclusion}

We have proposed a novel end-to-end learnable 3D object detector that exploits both LIDAR and cameras to perform very accurate 3D localization. Our approach uses continuous convolutions to fuse both sensor modalities at different levels of resolution by projecting the image features into bird's eye view. Our experimental evaluation on both KITTI \cite{kitti} and a large scale 3D object detection benchmark \cite{pixor} shows that our approach significantly outperforms the state of the art. 

% \clearpage

\bibliographystyle{splncs04}
\bibliography{egbib}
\end{document}